\documentclass{article} %
\usepackage{iclr2027_conference,times}

\usepackage{amsmath,amsfonts,bm}

\def\eqref#1{equation~\ref{#1}}

\def\1{\bm{1}}

\def\vmu{{\bm{\mu}}}

\def\vb{{\bm{b}}}
\def\vc{{\bm{c}}}

\def\vk{{\bm{k}}}
\def\vl{{\bm{l}}}

\def\vp{{\bm{p}}}

\def\vu{{\bm{u}}}

\def\vw{{\bm{w}}}
\def\vx{{\bm{x}}}
\def\vy{{\bm{y}}}
\def\vz{{\bm{z}}}

\def\mD{{\bm{D}}}

\def\mI{{\bm{I}}}

\def\mP{{\bm{P}}}
\def\mQ{{\bm{Q}}}

\def\mS{{\bm{S}}}

\def\mU{{\bm{U}}}

\def\mW{{\bm{W}}}

\DeclareMathAlphabet{\mathsfit}{\encodingdefault}{\sfdefault}{m}{sl}
\SetMathAlphabet{\mathsfit}{bold}{\encodingdefault}{\sfdefault}{bx}{n}

\newcommand{\E}{\mathbb{E}}

\newcommand{\R}{\mathbb{R}}

\usepackage[utf8]{inputenc}
\usepackage[T1]{fontenc}
\usepackage{hyperref}
\hypersetup{hidelinks}
\usepackage{url}
\usepackage{graphicx}
\usepackage{booktabs}
\usepackage{array}
\usepackage{multirow}
\usepackage{amsmath,amssymb}
\usepackage[table]{xcolor}
\usepackage{subcaption}
\usepackage{tikz}
\usetikzlibrary{positioning,arrows.meta,fit,backgrounds,calc,shapes.geometric}
\usepackage[capitalise,noabbrev]{cleveref}
\usepackage{wrapfig}
\usepackage{enumitem}

\newcommand{\ours}{DRAFT}

\newcommand{\threegrid}{$3{\times}3$}
\newcommand{\up}{$\uparrow$}
\newcommand{\down}{$\downarrow$}
\newcommand{\ci}[2]{{\scriptsize[#1, #2]}}
\newcommand{\best}[1]{\textbf{#1}}
\newcommand{\second}[1]{\underline{#1}}

\newcommand{\safefig}[2][\linewidth]{%
  \IfFileExists{#2}{\includegraphics[width=#1]{#2}}{%
    \fbox{\parbox[c][1.6in][c]{\dimexpr#1-2\fboxsep-2\fboxrule}{\centering\small Figure in preparation: \texttt{\detokenize{#2}}}}}}

\title{DRAFT: Decoding and Reconstructing\\3D Room Layouts from fMRI\\and Camera Trajectories}

\author{Chi-Wei Lee$^{1}$, An-Kai Chen$^{1}$, Cheng-Han Chiang$^{1}$, Yi-Chun Chang$^{2}$, Po-Chih Kuo$^{1}$\\
{\normalfont $^{1}$National Tsing Hua University \quad $^{2}$National Yang Ming Chiao Tung University}}

\iclrfinalcopy
\hypersetup{pdftitle={DRAFT: Decoding and Reconstructing 3D Room Layouts from fMRI and Camera Trajectories},
  pdfauthor={Chi-Wei Lee, An-Kai Chen, Cheng-Han Chiang, Yi-Chun Chang, Po-Chih Kuo}}
\begin{document}
\addtocontents{toc}{\protect\setcounter{tocdepth}{-1}}

\maketitle
\lhead{Preprint}
\suppressfloats[t]

\begin{abstract}
While functional magnetic resonance imaging (fMRI) decoders reconstruct images and 3D objects, recovering the metric geometry of rooms from a moving viewpoint remains an open challenge.
We introduce, to our knowledge for the first time, the task of reconstructing 3D layouts of held-out scenes using only the known camera trajectories and the fMRI signals of participants watching indoor walkthroughs, without any visual frames.
The task is hard because of the temporal mismatch between rapid visual changes and blurred fMRI signals.
Our method, \ours{} (\textbf{D}ecoding and \textbf{R}econstructing 3D room l\textbf{A}youts from \textbf{f}MRI and \textbf{T}rajectories), extracts a compact layout representation from the early visual cortex and fuses this neural code with the geometric free space derived from camera motion.
To evaluate it, we propose layout $r$, the Pearson correlation between predicted and ray-cast log depths.
On ten held-out scenes (20 participants), \ours{} reaches a layout $r$ of 0.480, outperforming the camera path alone (0.446) and seven fMRI decoders (at most 0.429).
\ours{} also works as a plug-in that improves all seven decoders without retraining, raising their layout $r$ from 0.171--0.429 to 0.487--0.493.
Brain activity and camera motion can thus recover the metric layout of unseen rooms, a first step toward reconstructing the 3D spaces people see.
\end{abstract}

\section{Introduction}
\label{sec:intro}

\begin{figure}[t]
  \centering
  \includegraphics[width=\linewidth]{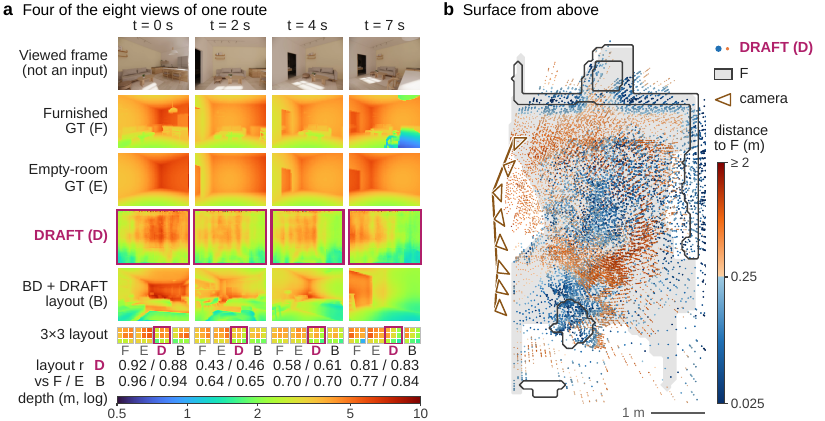}
  \caption{\textbf{\ours{} recovers where walls and floor are on a held-out route.}
\textbf{(a)} Four views of scene 4: frame (not an input), furnished (F) and empty-room (E) ground truth, \ours{} (D), Brain-Diffuser (BD) detail with the \ours{} layout (B), and layout $r$ of D and B against F and E.
\textbf{(b)} Merged surface of D from above, blue within 25\,cm of F.
Hand-picked route, ranked 60th in layout $r$ and 94th in F-score of 324 routes. Rule-chosen routes: \cref{fig:depth_atlas_1} and \cref{app:extended}.}
  \label{fig:teaser}
\end{figure}

\looseness=-1 Reconstructing a room needs more than a plausible picture. Walls must sit at the right distances, and views taken from different places must describe the same space.
Decoders of functional magnetic resonance imaging (fMRI) now reconstruct images and videos~\citep{ozcelik2023natural,scotti2024mindeye2,chen2023cinematic}, depth images~\citep{gaziv2021more,xia2024dream} and isolated 3D objects~\citep{gao2024mind3d}, yet to our knowledge none has recovered metric room geometry as the viewpoint moves.
We ask whether fMRI responses to indoor walkthrough videos support layout and surface reconstruction in scenes the model has never seen.

\looseness=-1 This is hard for three reasons. The viewpoint changes every second, the blood-oxygen-level-dependent (BOLD) signal of fMRI blurs several seconds together, and recordings are few.
So we do not decode depth maps directly. We decode a small layout code, 22 numbers per second that describe wall distances, floor extent, openings and depth range.
Ridge regressions, the readout, predict it from early visual cortex (EVC) and camera motion. EVC responses map the image point by point, so they carry image structure that correlates with depth. Scene-selective regions also respond to room layout (\cref{sec:related}), but in our data EVC predicted the code better (\cref{sec:complementary}).
The camera path also helps. Every visited position is free space, just as steps through a dark room show where the walls are not. We use it as a learned prior, not a bound on wall distance.

\looseness=-1 Our method \ours{} (\cref{fig:pipeline}) combines both and, like an architect's draft, fixes the coarse layout first and the details later.
It renders the code as coarse depth, and a small generator trained on real readout errors adds detail to eight consecutive views. \ours{} then replaces the broad structure (low spatial frequencies) of the views with a \threegrid{} \ours{} layout, predicted from the code and the camera path, and places the views in 3D at the known poses.

\looseness=-1 Across ten held-out scenes, the \ours{} layout's cell depths correlate with the true ones at 0.480 (layout $r$, \cref{sec:main}). This is above the best of seven adapted fMRI decoders (0.429), the fixed template (median training-scene depth, 0.324) and the generator alone (0.303).
Most of this comes from the camera path (0.446), but the brain adds to it. Adding the readout to the camera path raises layout $r$ in all ten scenes, and brain-input controls lower it in all ten (\cref{sec:complementary}).

\looseness=-1 The 3D surfaces tell a weaker story (\cref{sec:surface}). Because ridge regressions shrink the depth range, \ours{} ends with a range calibration, whose factor we fixed after earlier results on the same scenes.
Even so, unmodified MindEye and Brain-Diffuser have slightly better mean surface scores, and so does the generator alone with a milder calibration.
Guided by the \ours{} layout, these two image decoders reach the best fMRI-based mean surface scores, within noise of their unmodified versions.

\looseness=-1 We contribute a readout of a layout code from EVC and camera motion, and the \ours{} layout, which fuses this code with the camera path and raises the layout $r$ of all seven tested decoders without retraining (five post hoc). We also contribute a held-out-scene test of room geometry from fMRI, with ray-cast ground truth, fMRI-free and wrong-scene references and brain-input controls.

\section{Related Work}
\label{sec:related}

\paragraph{Image, depth and video reconstruction from fMRI.}
\looseness=-1 fMRI decoding reconstructs movies~\citep{nishimoto2011reconstructing} and images~\citep{shen2019deep} through learned feature mappings~\citep{kay2008identifying,naselaris2011encoding}.
The Natural Scenes Dataset (NSD)~\citep{allen2022massive} supports diffusion-based image decoding~\citep{takagi2023high,ozcelik2023natural,scotti2023reconstructing,scotti2024mindeye2}. MinD-Vis adds brain pretraining and MinD-Video extends it to video~\citep{chen2023seeing,chen2023cinematic}.
\citet{gaziv2021more} reconstruct depth and color images with monocular-depth supervision, and DREAM decodes depth, color and semantics for image reconstruction~\citep{xia2024dream}.
We instead test metric layout in held-out scenes against ray-cast geometry.

\paragraph{3D decoding and scene-selective cortex.}
\looseness=-1 MinD-3D reconstructs object meshes from fMRI of rotating objects~\citep{gao2024mind3d}, and Neuro-3D reconstructs colored object point clouds from electroencephalography (EEG)~\citep{guo2025neuro3d}. Closest to our setting, \citet{zeng2024geometry} decode the depth structure of first-person driving videos in a held-out virtual town by matching decoded embeddings to those of the town's own scenes. \ours{} instead predicts metric depth without the held-out scene's geometry and lifts it to 3D surfaces.
Scene-selective regions respond to spatial layout~\citep{epstein1998cortical,dilks2013occipital,henriksson2019rapid}, boundaries~\citep{park2011disentangling,julian2016occipital}, navigational affordances~\citep{bonner2017coding} and 3D surface configuration~\citep{lescroart2019human}, and support the cognitive map~\citep{epstein2017cognitive}.
For static photographs, Gabor-wavelet features predict their responses better than explicit surface models~\citep{shafer2026gabor}, and our readout also cannot separate depth from correlated image features.

\paragraph{Depth estimation and generative priors.}
\looseness=-1 Monocular depth models~\citep{eigen2014depth,ranftl2022towards,ranftl2021vision,yang2024depth,yang2024depthv2,bochkovskii2025depth,ke2024repurposing} now learn from large mixed datasets and diffusion priors.
Replacing low spatial frequencies resembles low-frequency guidance of diffusion models~\citep{choi2021ilvr} and spatial control of pretrained generators~\citep{zhang2023adding}. Our replacement acts on output depth without training, so any decoder can supply the detail. The camera path as a free-space prior relates to occupancy mapping~\citep{elfes1989using} and trajectory-based free-space prediction~\citep{wijmans2023emergence}.

\section{Task, Data and Evaluation}
\label{sec:setup}

\paragraph{Task.}
\looseness=-1 A participant watches a first-person walkthrough video of a furnished indoor scene during fMRI.
We split each video into \emph{routes} of eight consecutive one-second views. For each route, the model predicts eight depth maps and the 3D surface they form. Its inputs are the brain responses and the camera path (position, heading and pitch in each second), never the video frames.

\paragraph{Data.}
\looseness=-1 Twenty participants, 6 to 19 per scene, watched 80-second videos of ten furnished indoor scenes (\cref{tab:app_rooms}). Each scene was rendered in up to four interior styles that share the floor plan and a similar camera path but differ in materials, colors and furniture.
This gives 36 videos and 416 recordings (one participant watching one video). Each recording has 74 one-second samples, because the last six seconds are left for the response delay. Viewing was free, with no fixation task or eye tracking, and the same participants appear in training and held-out scenes.
Ground-truth radial depth (distance along each camera ray) is ray-cast from the 3D scene models at the recorded poses (90\textdegree{} horizontal field of view, $48\times64$ pixels). Nine non-overlapping routes per video give 324 routes.
Nuisance regression and z-scoring use whole scanner runs, which can mix training and held-out scenes but never a depth target (\cref{app:data}).

\paragraph{Protocol.}
\looseness=-1 We use leave-one-scene-out cross-validation over the ten scenes. Every model behind a test prediction is fitted without any style, recording or route of the held-out scene.
Models that take other models' predictions as input are fitted on out-of-fold predictions: for each training scene, models fitted without that scene predict it (\cref{app:impl}).
At test time, predictions are averaged over the recordings of the held-out video and over another style at matching poses, without depth labels (\cref{sec:pooling}). So results use pooled recordings, not single trials or single participants.

\paragraph{Comparisons.}
\looseness=-1 We adapt seven fMRI decoders to our folds. Five are published: image and depth decoding~\citep{takagi2023high,takagi2023improving}, the DREAM depth branch~\citep{xia2024dream}, Brain-Diffuser~\citep{ozcelik2023natural} and MindEye~\citep{scotti2023reconstructing}. Two are simple ridge decoders: linear depth decoding and retrieval of CLIP (Contrastive Language-Image Pre-training) image embeddings~\citep{radford2021learning}, like the retrieval branch of MindEye. The published decoders keep their released components and original brain regions. Depth Anything V2 (DA V2)~\citep{yang2024depthv2}, with its scale calibrated on the training scenes, turns decoded images into radial depth. The decoders may read scene-selective regions and use large pretrained models, but they see no camera input and decode each view separately. Trained on nine scenes, Brain-Diffuser and MindEye identify the viewed frame only weakly (\cref{app:baselines}). We also use two fMRI-free references. The \emph{fixed template} is the median training-scene depth, and the per-second median is this median taken separately at each video second (\cref{app:floors}).

\paragraph{Metrics.}
\looseness=-1 Our main metric, \textbf{layout $r$}, asks whether a prediction knows which parts of a view are near or far. It is the Pearson correlation between the predicted and true median log depths of the nine cells of a \threegrid{} grid over the view.
\textbf{Row order} is the fraction of correctly ordered cell pairs in the same row, and \textbf{left/right} the fraction of views whose deeper half is identified. \textbf{Pixel MAE} (mean absolute error) is the depth error in meters over valid pixels. All four are averaged over the 324 routes (\cref{app:metric_definitions}).
For surfaces, we back-project each view along its camera rays at the known pose. The eight views of a route then merge on 5\,cm voxels into one point set, the surface, without extracting a mesh. We compare it with the ground-truth points visible from the same poses, not with the whole room mesh.
The \textbf{F-score} (F@25cm) is the harmonic mean of \textbf{precision}, the fraction of predicted points within 25\,cm of a true point, and \textbf{recall}, the fraction of true points within 25\,cm of a predicted point. The \textbf{Chamfer distance} averages the nearest-point distances in both directions, in meters.
Surface metrics and paired contrasts weight scenes equally, with 95\% confidence intervals (CIs) from 20{,}000 paired scene-bootstrap resamples and the number of scenes that favor each method. Intervals are not corrected for multiple comparisons and can be too narrow, as scenes share participants and some scanner runs. So four main layout contrasts, named after the results were known, also get sign-flip $p$ values with Holm correction. Some other contrasts get uncorrected $p$ values. An interval that includes zero does not prove two methods equal (\cref{app:stats}).

\section{Method}
\label{sec:method}

\begin{figure}[t]
  \centering
  \includegraphics[width=\linewidth]{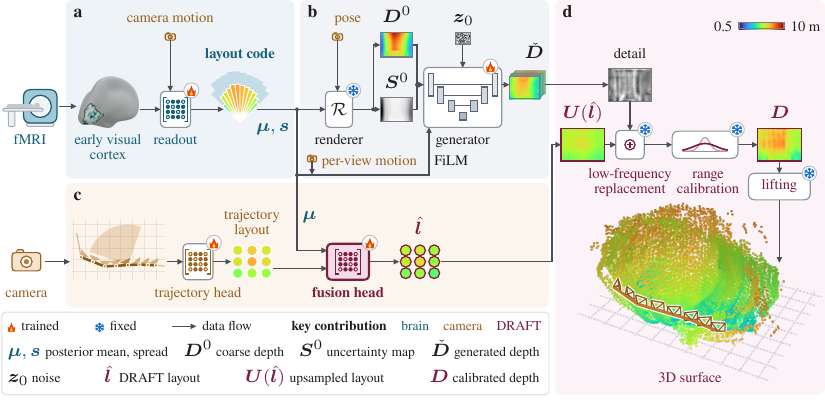}
  \caption{\textbf{Overview of \ours{}: the coarse layout is fixed before the detail.}
(a) The readout maps EVC and camera motion to the layout code, which (b) a renderer and generator turn into eight depth maps.
(c) Trajectory and fusion heads give the \ours{} layout, which (d) sets the broad structure of any depth map before range calibration and lifting to 3D.
Real data of the \cref{fig:teaser} route, except the noise $\vz_0$.}
  \label{fig:pipeline}
\end{figure}

\subsection{Readout of the layout code}
\label{sec:readout}
\paragraph{Layout code.}
\looseness=-1 The code $\vy(\mD)\in\R^{22}$ summarizes one view's depth map $\mD$ in log radial depth,
\begin{equation}
  \vy(\mD)=\big[\,w_1,\dots,w_{12},\ f_1,\dots,f_4,\ o_1,\dots,o_4,\ \operatorname{med}\log\mD,\ \operatorname{IQR}\log\mD\,\big].
  \label{eq:code}
\end{equation}
Here $w_k$ is the wall distance near the horizon in azimuth bin $k$, $f_b$ the floor extent in vertical band $b$, $o_q$ the share of far pixels in quadrant $q$ and IQR the interquartile range (\cref{app:impl}).

\paragraph{Readout.}
\looseness=-1 Six ridge regressions map EVC responses and 12 camera-motion regressors to the code $\vy$, standardized on the training scenes. They differ in voxel count, ridge penalty and response window, and their mean is the readout $\hat\vy_t$ (\cref{eq:readout}). EVC comprises V1, V2, ventral V3 (V3v) and human V4 (hV4), whose most reliable voxels are reduced to 16 principal components. The regressors are a constant, speed and turning speed at lags of 0 to 4\,s, and a linear trend. The regressions are fitted to participant-averaged responses of the nine training scenes and applied to each held-out recording.

\paragraph{Pooling across styles.}\label{sec:pooling}
\looseness=-1 One recording is noisy, and other styles of a scene often pass the same poses. So each second we average the readout of the evaluated recording with the participant-averaged readout of at most one other style, weighted by recording counts. Pooling never uses depth labels or pairs different seconds, but furniture may differ across styles (\cref{app:pooling_diagnostics}). The pooled readout of each code entry $j$ is its posterior mean $\mu_j$. Because readout reliability varies, each entry also gets a posterior spread $s_j$, its out-of-fold residual standard deviation (\cref{app:impl}). Together they form a point estimate with an error bar, not a full Bayesian posterior.

\subsection{Coarse depth and flow-matching generator}
\label{sec:generator}
\paragraph{Renderer.}\label{sec:renderer}
\looseness=-1 A renderer turns the posterior mean and pose into coarse inverse depth $\mD^{0}_t$. This map holds walls, eye height, floor extents and openings, matched to the code's median and IQR.
A first-order approximation carries the posterior spreads into an uncertainty map. It follows only each pixel's dominant wall distance, not the renderer's full covariance (\cref{app:impl}).

\paragraph{Generator.}
\looseness=-1 Coarse depth has walls and floors but no detail, so a generator adds it, modeling the eight views of a route jointly. It learns the residual between the true inverse depth $\mQ_{\rm GT}$ and $\mD^{0}$ with rectified flow~\citep{liu2023flow}, a form of flow matching~\citep{lipman2023flow}. A penalty asks the generated depth to agree with the decoded layout code, most strongly where the code is reliable,
\begin{equation}
  \mathcal{L}=\E\,\big\|v_\theta(\vz_\tau,\tau,\vc)-(\vz_1-\vz_0)\big\|^2
  +\lambda\,\operatorname{mean}\!\left[\widetilde{\vw}\left(\mathcal{A}_i(\hat{\mQ}_1)-\vmu\right)^2\right],
  \qquad \vz_\tau=(1-\tau)\vz_0+\tau\vz_1.
  \label{eq:flow}
\end{equation}
Here $\tau\in[0,1]$ is the flow time, $v_\theta$ the velocity of a network with weights $\theta$, $\vz_1=(\mQ_{\rm GT}-\mD^0)/\rho$ the residual scaled by a training-scene constant $\rho$, $\vz_0$ Gaussian noise and $\vc$ the conditions (coarse depth, uncertainty map, posterior and motion). The penalty acts on the predicted depth $\hat\mQ_1=\mD^0+\rho\,\big(\vz_\tau+(1-\tau)v_\theta\big)$.
A fixed affine inverse of the renderer, $\mathcal A_i$, returns the code implied by a depth map, standardized like the posterior mean of the sampled recording $i$. The weights $\widetilde\vw$ are the normalized inverse variances $s_j^{-2}$, and $\lambda=0.1$ (\cref{app:flow_terms}). Without the penalty, the generator alone has a higher layout $r$ (0.418 against 0.301, ten-scene means), but \ours{} has worse surfaces (\cref{app:no_aux_ablation}).
The network is a 3D U-Net~\citep{cicek2016unet} over space and views, conditioned by feature-wise linear modulation (FiLM)~\citep{perez2018film}.
Training pairs each route's ray-cast depth with the out-of-fold posterior of a random recording of its video, so the conditions contain real readout errors. At test time, $\check{\mD}_t$ inverts the mean inverse-depth sample over the video's recordings.

\paragraph{Loss correction.}\looseness=-1 An earlier implementation compared the two sides of the penalty in different coordinates, so all main results use ten generators retrained with the corrected loss (\cref{app:loss_audit,app:corrected_pilot}). The correction lowers the layout $r$ of the generator alone from 0.369 to 0.303. The \ours{} layout barely changes (0.486 to 0.480), so part of the fusion gain reflects a weaker generator.

\subsection{Trajectory head, fusion and range calibration}
\label{sec:fusion}
\paragraph{Trajectory head.}\label{sec:tfs}
\looseness=-1 The camera path also carries layout information. In each of 12 directions, a free-space feature is the largest distance from the camera to a position visited within a few seconds, in that direction (\cref{eq:tfs}).
The trajectory head, a ridge regression on these and other path features, predicts a \threegrid{} log-depth layout, the trajectory layout (\cref{eq:fusion}).

\paragraph{Fusion.}
\looseness=-1 The code and the trajectory layout $\vl^{\rm T}_t$ make different errors, so a second ridge regression, the fusion head, combines them as a stacking model~\citep{wolpert1992stacked},
\begin{equation}
  \hat\vl_t=\mW_{\rm F}^\top\big[\vmu_t;\ \vl^{\rm T}_t\big]+\vb_{\rm F},\qquad \vl^{\rm T}_t=\mW_{\rm T}^\top\boldsymbol\phi_t+\vb_{\rm T},
  \label{eq:fusion}
\end{equation}
where $\boldsymbol\phi_t$ holds the 111 path features of the trajectory head. The fusion head is fitted on out-of-fold inputs, and inner cross-validation on the same inputs, without further nesting, chooses its penalty (\cref{app:impl}). Its output $\hat\vl_t\in\R^{9}$ is the \ours{} layout, which sets the low spatial frequencies of the log depth. The generator supplies the detail. This low-frequency replacement has no fitted parameters, so it applies to any other decoder's depth as well (\cref{sec:baselines}),
\begin{equation}
  \log\mD_t = G_\sigma * \mU(\hat\vl_t) + \big(\log\check{\mD}_t - G_\sigma * \log\check{\mD}_t\big),
  \qquad \sigma=4\ \text{pixels},
  \label{eq:lowfreq}
\end{equation}
where $\mU$ is bilinear upsampling to $48\times64$ and $G_\sigma *$ Gaussian low-pass filtering. Layout $r$ scores these low frequencies, so it depends mainly on the \ours{} layout, not on the detail (\cref{app:floors}).

\paragraph{Range calibration.}
\looseness=-1 Ridge regressions shrink predictions toward the mean, so the cell log depths of the \ours{} layout have only 0.49 of the true spread (generator 0.78). Before lifting, we therefore map $\log\mD_t$ to $m+a(\log\mD_t-m)$, with $m$ the video's mean predicted log depth, which uses no test depth. This map keeps layout $r$, row order and left/right except where depth is clipped. Inner validation on each fold's training scenes selects $a=2$ in every fold by Chamfer distance, or 2 or 2.5 by F-score (\cref{app:range_sensitivity}). We added this step and its selection after earlier results on these scenes.

\section{Experiments}
\label{sec:experiments}

\begin{table}[t]
  \centering
  \caption{\textbf{Main results: fusion gives the best layout at no measurable cost in surface accuracy.} Each layout replaces the broad structure of one generator's depth. C: EVC, M: camera motion (12 readout regressors, 72 trajectory-head features), T: free space. Row: row order, L/R: left/right, bold: best, underline: second. Layout, MAE: 324-route means, others: ten-scene means. Last rows: \ours{} minus generator alone ($^\ast$: 95\% paired scene-bootstrap CI excludes zero).}
  \label{tab:main}
  \footnotesize
  \setlength{\tabcolsep}{5pt}
  \def\dg#1{{\scriptsize#1}}\def\sa{\rlap{\smash{$^{\textstyle\ast}$}}}
  \begin{tabular}{@{}ll rrr rrr@{}}
    \toprule
    & & \multicolumn{3}{c}{Layout} & \multicolumn{1}{c}{Pixel} & \multicolumn{2}{c}{Surface} \\
    \cmidrule(lr){3-5}\cmidrule(lr){6-6}\cmidrule(l){7-8}
    Layout from & Inputs & $r$\,\up & Row\,\up & L/R\,\up & MAE (m)\,\down & F@25cm\,\up & Chamfer (m)\,\down \\
    \midrule
    Generator alone & none & 0.303 & 0.626 & 0.602 & 1.507 & 0.478 & 0.449 \\
    Readout, \threegrid{} target & C, M & 0.440 & 0.656 & 0.621 & \second{1.176} & \best{0.485} & \second{0.417} \\
    Trajectory head & M & 0.390 & 0.665 & 0.632 & 1.264 & 0.447 & 0.443 \\
    Trajectory head & M, T & \second{0.446} & \second{0.688} & \second{0.658} & 1.252 & 0.465 & 0.427 \\
    Fusion head (\ours{}) & C, M, T & \best{0.480} & \best{0.696} & \best{0.668} & \best{1.153} & \second{0.484} & \best{0.405} \\
    \midrule
    \multicolumn{2}{@{}l}{\ours{} minus generator alone} & \dg{+0.172\sa} & \dg{+0.066\sa} & \dg{+0.065\sa} & \dg{\textminus0.356\sa} & \dg{+0.006} & \dg{\textminus0.044\sa} \\
    \multicolumn{2}{@{}l}{Scenes with \ours{} better} & \dg{10/10} & \dg{10/10} & \dg{8/10} & \dg{10/10} & \dg{4/10} & \dg{8/10} \\
    \bottomrule
  \end{tabular}
\end{table}

\begin{table}[t]
  \centering
  \caption{\textbf{Comparison with fMRI decoders: every decoder scores best with the \ours{} layout, and none matches it given only the camera path.} Alone: unmodified. + path, + \ours{}: with the trajectory layout (camera path only) or the \ours{} layout, post hoc except for linear depth and Brain-Diffuser. Averaging as in \cref{tab:main}. Bold: row best. $^\ast$: 95\% CI of the gain over alone excludes zero (\cref{tab:baselines_ci}). Unmodified MindEye and Brain-Diffuser keep slightly higher F-scores than \ours{}. Decoders: 418 recordings, \ours{}: 416. $^\ddagger$DA V2 on the frames. $^\dagger$Our adaptation (\cref{app:dream_adapter}).}
  \label{tab:baselines}
  \fontsize{8}{8.6}\selectfont
  \setlength{\tabcolsep}{3.4pt}
  \def\sg{$^\ast$}\def\ns{\phantom{$^\ast$}}\def\dh{+\,\ours{}}
  \begin{tabular}{@{}l ccc@{\hspace{13pt}}ccc@{\hspace{13pt}}ccc@{}}
    \toprule
    & \multicolumn{3}{c}{Layout $r$\,\up} & \multicolumn{3}{c}{F@25cm\,\up} & \multicolumn{3}{c}{Chamfer (m)\,\down} \\
    \cmidrule(r{6pt}){2-4}\cmidrule(lr{6pt}){5-7}\cmidrule(l){8-10}
    Method & alone & + path & \dh & alone & + path & \dh & alone & + path & \dh \\
    \midrule
    Linear depth & 0.429 & 0.454 & \best{0.490}\sg & 0.393 & 0.440 & \best{0.455}\sg & 0.508 & 0.454 & \best{0.436}\sg \\
    CLIP retrieval & 0.261 & 0.451 & \best{0.488}\sg & 0.455 & 0.473 & \best{0.493}\sg & 0.421 & 0.406 & \best{0.383}\sg \\
    Image decoding & 0.171 & 0.450 & \best{0.488}\sg & 0.442 & 0.456 & \best{0.474}\sg & 0.446 & 0.426 & \best{0.403}\sg \\
    Depth decoding & 0.322 & 0.455 & \best{0.491}\sg & 0.317 & 0.435 & \best{0.449}\sg & 0.727 & 0.462 & \best{0.444}\sg \\
    DREAM depth branch$^\dagger$ & 0.428 & 0.450 & \best{0.488}\sg & 0.368 & 0.435 & \best{0.449}\sg & 0.561 & 0.464 & \best{0.445}\sg \\
    Brain-Diffuser & 0.324 & 0.458 & \best{0.493}\sg & 0.492 & 0.476 & \best{0.499}\ns & 0.389 & 0.400 & \best{0.377}\ns \\
    MindEye & 0.270 & 0.451 & \best{0.487}\sg & 0.494 & 0.476 & \best{0.499}\ns & 0.381 & 0.406 & \best{0.376}\ns \\
    \multicolumn{3}{@{}l}{\emph{Best with the \ours{} layout (with $^\ast$)}} & 7/7 (7) & & & 7/7 (5) & & & 7/7 (5) \\
    \midrule
    \ours{} (own generator) & -- & -- & 0.480\ns & -- & -- & 0.484\ns & -- & -- & 0.405\ns \\
    Camera path alone & -- & 0.446 & --\ns & -- & 0.465 & --\ns & -- & 0.427 & --\ns \\
    Fixed template & 0.324 & -- & --\ns & 0.391 & -- & --\ns & 0.524 & -- & --\ns \\
    Video frames + DA V2$^\ddagger$ & 0.865 & -- & --\ns & 0.742 & -- & --\ns & 0.196 & -- & --\ns \\
    \bottomrule
  \end{tabular}
\end{table}

\subsection{Layout accuracy}
\label{sec:main}
\looseness=-1 \textbf{The \ours{} layout is more accurate than every fMRI decoder that does not use it, even with the camera path.}
Its layout $r$ is higher than that of linear depth in seven scenes and of MindEye and Brain-Diffuser in all ten (no sign-flip test, \cref{tab:baselines_ci}). Unlike these decoders, the fusion head is fitted to the cell depths that layout $r$ scores. Given the camera path through the trajectory layout, each decoder still trails \ours{} by 0.024 to 0.032 (ten-scene means). It trails in eight to ten scenes (sign-flip $p\le0.010$), and this comparison is post hoc for five decoders (\cref{tab:baselines_ci}).

\begin{figure}[t]
 \centering
 \includegraphics[width=\linewidth]{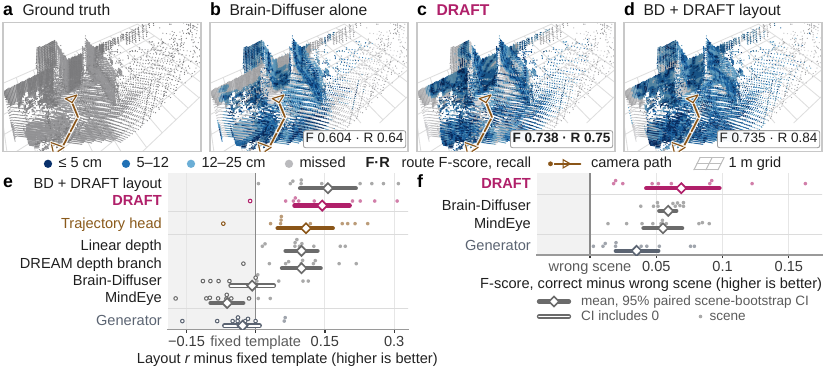}
 \caption{\textbf{Surface scores separate methods less than layout scores do.}
(a to d) Display route (scene 6), one of the top five for \ours{} among routes where \ours{} and BD + \ours{} layout beat unmodified Brain-Diffuser (BD) in F-score. b to d: true points below 2.4\,m by distance to the nearest prediction (F: F-score, R: recall, \cref{fig:oblique_routes}). On rule-chosen routes, \ours{} has the lowest F-score of the generator, \ours{} and BD + \ours{} layout in seven of ten (\cref{tab:atlas_routes}).
(e) Layout $r$ gain over the fixed template. (f)~Post hoc F-score gain over a wrong scene.}
 \label{fig:surface_composite}
\end{figure}

\looseness=-1 \textbf{The gain is more than a good guess about indoor rooms.}
Indoor views share a common shape, floor near and walls farther away, so a template gets part of the layout right. The \ours{} layout beats the fixed template in nine scenes and the per-second median (layout $r$ 0.241) in all ten. Brain-Diffuser, MindEye and the generator alone do not beat the fixed template (\cref{fig:surface_composite}e).

\begin{table}[t]
  \centering
  \caption{\textbf{Layout $r$ gains of \ours{} hold in nine or ten scenes, and surface gains are smaller.} Ten-scene means of paired differences (positive: first better), 95\% paired scene-bootstrap CI, scenes favoring the first. $r$: layout $r$, BD: Brain-Diffuser, M, T: as in \cref{tab:main}, cosine: similarity of centered codes. MAE, Chamfer in m. Sign-flip $p$: $^{\rm a}$0.002, $^{\rm b}$0.004, $^{\rm c}$0.012, $^{\rm d}$0.025, others untested.}
  \label{tab:contrasts}
  \scriptsize
  \setlength{\tabcolsep}{1.7pt}
  \def\ci#1#2{\,[#1,\,#2]}\def\mn{\textminus}\def\sa{$^{\rm a}$}\def\saa{$^{\rm b}$}\def\ssec{$^{\rm c}$}\def\spar{$^{\rm d}$}
  \begin{minipage}[t]{0.497\linewidth}
  \begin{tabular}[t]{@{}llrr@{}}
    \toprule
    Comparison & Metric & Gain [95\% interval] & Scenes \\
    \midrule
    \ours{} vs generator & $r$ & +0.172 \ci{0.135}{0.210} & 10\sa \\
     & Row & +0.066 \ci{0.040}{0.094} & 10 \\
     & L/R & +0.065 \ci{0.028}{0.101} & 8 \\
     & MAE & +0.356 \ci{0.304}{0.408} & 10 \\
    \quad vs fixed template & $r$ & +0.144 \ci{0.084}{0.204} & 9\saa \\
    \quad vs per-second median & $r$ & +0.225 \ci{0.149}{0.298} & 10\sa \\
    BD vs fixed template & $r$ & \mn0.008 \ci{\mn0.055}{0.039} & 5 \\
    MindEye vs fixed template & $r$ & \mn0.062 \ci{\mn0.098}{\mn0.027} & 2 \\
    Generator vs fixed template & $r$ & \mn0.028 \ci{\mn0.068}{0.009} & 2 \\
    \midrule
    Readout vs trajectory (M) & $r$ & +0.050 \ci{0.015}{0.092} & 8\spar \\
    Trajectory (M, T) vs (M) & $r$ & +0.050 \ci{0.022}{0.076} & 8\ssec \\
    \ours{} vs trajectory head & $r$ & +0.035 \ci{0.025}{0.046} & 10\sa \\
     & MAE & +0.103 \ci{0.068}{0.133} & 9 \\
    \quad on BD depth & $r$ & +0.036 \ci{0.025}{0.048} & 10\sa \\
    \quad on linear depth & $r$ & +0.037 \ci{0.026}{0.048} & 10\sa \\
    \bottomrule
  \end{tabular}
  \end{minipage}\hfill
  \begin{minipage}[t]{0.497\linewidth}
  \begin{tabular}[t]{@{}llrr@{}}
    \toprule
    Comparison & Metric & Gain [95\% interval] & Scenes \\
    \midrule
    Readout vs motion alone & $r$ & +0.098 \ci{0.053}{0.151} & 9\saa \\
    \ours{} vs readout & Row & +0.040 \ci{0.023}{0.057} & 10 \\
    EVC vs motion alone & Cosine & +0.219 \ci{0.150}{0.290} & 10\sa \\
    EVC vs scene-selective & Cosine & +0.160 \ci{0.102}{0.226} & 9 \\
    \ours{} vs replaced EVC & $r$ & +0.037 \ci{0.025}{0.051} & 10\sa \\
     & F-score & +0.036 \ci{0.024}{0.050} & 10\sa \\
    \quad vs shifted BOLD & F-score & +0.044 \ci{0.029}{0.060} & 10\sa \\
    \midrule
    \ours{} vs generator & Chamfer & +0.044 \ci{0.018}{0.069} & 8 \\
     & Precision & +0.055 \ci{0.041}{0.071} & 10 \\
     & F-score & +0.006 \ci{\mn0.013}{0.026} & 4 \\
    \ours{} vs no detail & F-score & +0.042 \ci{0.036}{0.047} & 10 \\
     & Chamfer & +0.049 \ci{0.040}{0.057} & 10 \\
    \ours{} vs wrong scene & F-score & +0.069 \ci{0.042}{0.098} & 10 \\
    \quad minus BD's gain & F-score & +0.010 \ci{\mn0.019}{0.042} & 5 \\
    \bottomrule
  \end{tabular}
  \end{minipage}
\end{table}
\begin{figure}[t]
 \centering
 \includegraphics[width=\linewidth]{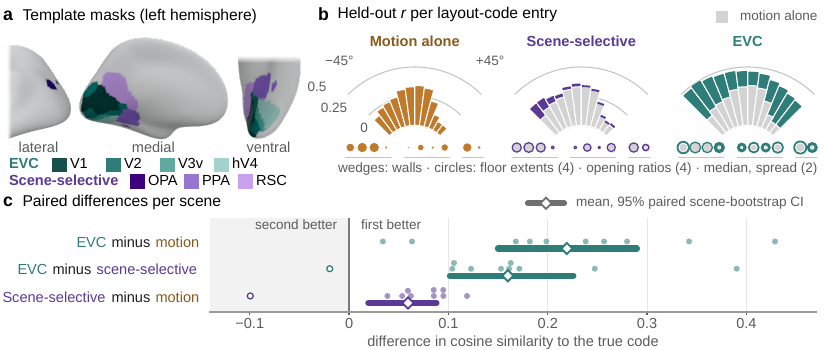}
 \caption{\textbf{EVC predicts the layout code best of the tested regions.}
\textbf{(a)} Template masks, not individual localizers.
\textbf{(b)} Held-out $r$ per layout-code entry (ten-scene mean), from motion alone or with 1{,}000 voxels of one mask.
\textbf{(c)} Per-scene differences in cosine similarity to the true code.}
 \label{fig:roi_evidence}
\end{figure}

\subsection{Camera path and brain contributions}
\label{sec:complementary}
\looseness=-1 \textbf{The camera path gives most of the layout, and the brain adds to it in every scene.}
From the camera path alone, the trajectory head reaches a layout $r$ of 0.446 (\cref{tab:main}). To score the readout by layout $r$ directly, we also fit it to the \threegrid{} cell depths. Adding EVC to its 12 camera-motion regressors raises layout $r$ from 0.338 to 0.440 in nine scenes (\cref{tab:contrasts}). Fusing the code with the trajectory layout raises layout $r$ by 0.035 in all ten scenes and lowers pixel MAE in nine.

\looseness=-1 \textbf{Two brain-input controls separate the brain's gain from camera motion.}
Because the readout also uses camera-motion regressors, the fusion gain does not isolate the brain. Both controls keep these regressors and lower the layout $r$ and F-score of \ours{} in all ten scenes.
Replacing the EVC features (16 principal-component scores) with their training-scene mean (replaced EVC) lowers them by 0.037 and 0.036 (\cref{app:feature_zero}).
Circularly shifting each recording's BOLD series by 37 of 80 seconds (shifted BOLD) and refitting the readout lowers them by 0.043 and 0.044 (one fixed shift, not a permutation test, \cref{app:circular_corrected}). At the readout level, the EVC gain peaks at a lag of 3 to 4\,s and beats all 20 circular shifts (\cref{app:roi_readout}). Post hoc, the brain's gain persists in single recordings (0.012, eight scenes) and in all 300 tested pipeline settings. It is no larger where the path covers less of the view, and more recordings would add only about 0.002 (\cref{app:posthoc}).

\looseness=-1 \textbf{EVC predicts the code best among the tested regions.}
With 1{,}000 voxels per mask and the camera-motion regressors, EVC predicts the code better than motion alone in all ten scenes. It also beats the scene-selective occipital place area (OPA), parahippocampal place area (PPA) and retrosplenial complex (RSC) together in nine (\cref{fig:roi_evidence}, \cref{app:roi_readout}).
Without functional localizers or dorsal motion areas, this only ranks the masks and shows no biological cause.

\subsection{Guiding other decoders and agreement across views}
\label{sec:baselines}
\looseness=-1 \textbf{Other decoders inherit the \ours{} layout.}
All seven decoders score best with it in all three metrics of \cref{tab:baselines}, with a 95\% CI excluding zero in 17 of 21 cases, all but the surfaces of Brain-Diffuser and MindEye. For Brain-Diffuser, it raises layout $r$ from 0.324 to 0.493 in all ten scenes, the highest fMRI-based value. The other six reach about 0.49 (five post hoc), above their versions with the trajectory layout in all ten scenes (\cref{app:transfer}). Layout $r$ then mostly scores the \ours{} layout (\cref{sec:fusion}), so surfaces are the real test.

\looseness=-1 \textbf{The \ours{} layout also makes the views agree.}\label{sec:horizon}
The \ours{} layout lowers the change in 3D error between views 1\,s apart for every scored depth source. Yet this measure also rewards views that are wrong in the same way (\cref{tab:horizon_ci,app:horizon}).
\ours{} views agree more than Brain-Diffuser's with the \ours{} layout (mean change 0.63 against 0.81\,m, all ten scenes, \cref{app:horizon}). But linear depth with the same layout, also decoded view by view, agrees as well or better (0.58\,m).

\subsection{Eight-view 3D surfaces}
\label{sec:surface}
\looseness=-1 \textbf{After range calibration, the \ours{} layout costs no measurable surface accuracy.}
Calibration brings the cell log depths of \ours{} to 0.97 of the true spread. \ours{} has a lower Chamfer distance than the generator alone (eight scenes), a higher precision (all ten) and a similar F-score (\cref{tab:contrasts}). But its Chamfer distance is not lower than the generator's at its own selected factor (1.5, \cref{tab:nested_a}). The readout with a \threegrid{} target, without free space, reaches a similar F-score (0.485).
Without calibration, fusion lowers the F-score of the generator alone from 0.477 to 0.416 and raises its Chamfer distance by 0.065\,m in all ten scenes. The \ours{} layout also lowers the uncalibrated F-score of the three decoders with the best unmodified surfaces, because ridge heads shrink the depth range (\cref{tab:main_uncal,app:transfer}).

\looseness=-1 \textbf{The calibration factor was fixed after earlier results on these scenes.}
Selecting the factor by F-score instead of Chamfer distance picks 2 or 2.5 and lowers the held-out F-score of \ours{} to 0.477. At their Chamfer-selected factors, \ours{}, the generator alone, Brain-Diffuser and MindEye differ in F-score by at most 0.016, the generator alone slightly ahead, within noise (\cref{tab:nested_a}).

\looseness=-1 \textbf{Unmodified MindEye and Brain-Diffuser reach slightly higher mean F-scores than \ours{}} (0.494 and 0.492 against 0.484) and lower Chamfer distances (\cref{tab:baselines}). Only the lower Chamfer distance of MindEye holds beyond noise (nine scenes, \cref{app:floors}).
Their detail comes from large pretrained image models, ours from a small generator trained on nine scenes. Removing our detail lowers the F-score by 0.042 in all ten scenes (\cref{tab:contrasts}), but this detail is mostly one pattern shared by all scenes and correlates with the true detail at only 0.026 (\cref{app:floors}). So scene-specific geometry comes mainly from the \ours{} layout.
Against decoders with the trajectory layout, \ours{} is within noise of CLIP retrieval, Brain-Diffuser and MindEye in F-score. It is ahead of the other four, each in at least eight scenes (post hoc sign-flip tests, \cref{tab:baselines_ci}).
With the \ours{} layout, Brain-Diffuser's detail raises recall over \ours{} and slightly lowers precision, in all ten scenes. This gives the best fMRI-based mean F-score (0.499), within noise of both unmodified decoders (\cref{app:layout_guidance}). MindEye with the \ours{} layout ties it post hoc and has the lowest Chamfer distance (0.376\,m).

\looseness=-1 \textbf{Surface scores separate the methods little, for two reasons.} First, even a wrong scene scores well. Depth predicted for another scene and lifted at the evaluated poses (post hoc) keeps most of the F-score of \ours{}, Brain-Diffuser and MindEye. The correct scene adds only 0.055 to 0.069, in all ten scenes each (\cref{tab:floors}, \cref{fig:surface_composite}f). Subtracting this wrong-scene score does not favor \ours{} over MindEye (post hoc, \cref{app:posthoc}). Second, the scores reward depth range. DA V2 gives decoded images nearly the true range, 0.84 of the true spread against 0.46 to 0.54 for our ridge layouts (\cref{app:dream_adapter}). This raises recall even for a wrong layout.

\looseness=-1 \textbf{Once depth scale is removed, \ours{} has lower depth error than the unmodified image decoders, but this is a low bar.}
Scored without range calibration for every method, \ours{} has a lower scale-invariant depth error~\citep[SI-log,][]{eigen2014depth} than the trajectory head, the generator, Brain-Diffuser and MindEye in all ten scenes (sign-flip $p=0.002$). After post hoc median alignment of each view to the true depth, its F-score beats these four in nine or ten scenes each.
Yet Brain-Diffuser and MindEye have a higher SI-log than the fixed template (57.3 and 59.5 against 52.8, and 49.6 for \ours{}). Linear depth ties \ours{}, and Brain-Diffuser with the \ours{} layout nearly ties it in SI-log and beats it after alignment. The fixed template, Brain-Diffuser and MindEye also have more pixels within a factor of 1.25 of the true depth (\cref{app:aligned,tab:aligned}).

\section{Discussion and Conclusion}
\label{sec:conclusion}

\looseness=-1 To our knowledge, \ours{} is the first method to decode from fMRI the metric layout of rooms seen from a moving viewpoint and to test it against ray-cast ground truth in held-out scenes. It reads a layout code each second from EVC and camera motion, fuses it with the camera path and turns it into depth maps and 3D surfaces. In ten held-out scenes, its layout $r$ is 0.480, above seven adapted fMRI decoders (at most 0.429) and the camera path alone (0.446), and brain-input controls lower it in every scene. Used as a plug-in prior without retraining, the \ours{} layout raises the layout $r$ of all seven decoders to about 0.49 (five post hoc) and gives Brain-Diffuser and MindEye the best fMRI-based mean surface scores. Its fMRI-free and wrong-scene references and brain-input controls separate what the camera path, the brain and pretrained image models contribute.

\looseness=-1 \textbf{Limitations.} \ours{} needs known camera paths and pooled recordings, was developed on these scenes with several analyses post hoc, and its brain gain was not measurable beyond low-level image features and replicated within only two of five withheld scenes (\cref{app:lowlevel,app:lockbox}).

\looseness=-1 \textbf{Outlook.} Designed routes, more participants and estimated camera paths can test single-trial decoding in new scenes, and generative 3D scene models could turn the \ours{} layout into complete room reconstructions.\label{page:last_main}

\clearpage
\subsection*{AI use statement}
We used large language model assistants extensively, under the authors' supervision. They helped write and debug the analysis, model and evaluation code, develop the method, and propose, refine, preregister, run and audit experiments and analyses. They also helped interpret results, search the literature and verify references, translate the authors' notes into English, and draft and edit the text, equations and figures.
They were not used to collect, generate or modify the fMRI data or the ground-truth geometry, no synthetic data enter any reported result, and the paper makes no theorem-style mathematical claims.
The authors chose the research questions, checked all analyses, numbers, references and claims, and take full responsibility for this paper, including text, results and artifacts made with these tools.

\subsection*{Ethics statement}
This work analyzes existing fMRI recordings. The original study was approved by the institutional review board of the authors' institution, and all participants gave written informed consent.
Analyses use coded participant identifiers and concern brain responses to rendered indoor scenes.
The method depends on participant cooperation, known camera paths and repeated recordings, and it has not been tested for unconstrained or covert decoding. Neural data and derived readouts are sensitive.

\subsection*{Reproducibility statement}
\Cref{sec:setup} and \cref{app:data} describe the data, the cross-validation protocol and the metrics.
\Cref{sec:method} and \cref{app:impl} give every model component with its hyperparameters, and \cref{app:stats} describes the statistical procedure.
Every fitted parameter and every hyperparameter chosen by inner validation is selected without the held-out scene, and each model uses one fixed seed. Fixed design settings were chosen during development, which used all ten scenes. The clearest case is the range-calibration factor $a=2$ (\cref{sec:fusion}), fixed after earlier results on the same scenes (\cref{app:range_sensitivity}). Nested selection by Chamfer distance picks the same value in all ten folds, and selection by F-score picks 2 or 2.5 (\cref{app:floors}).
Decoder predictions were frozen before scoring under a written plan (\cref{app:baselines}), and the transfer of the layout to five of the seven decoders is a post hoc analysis (\cref{app:transfer}). One generator fold trains in about 15 minutes on one GPU (\cref{app:impl}).
We will also release a static demo website that runs offline, with an interactive viewer of the ten atlas routes, one per scene, chosen by a fixed rule, which shows predicted depth maps and merged 3D surfaces next to the ground truth.
Upon publication, we will release the code for the method, trained models, evaluation manifests and derived readouts. Voxel-level fMRI data remain subject to the original consent and institutional data-sharing rules, and the derived readouts alone are not sufficient to retrain the cortical decoder.

\bibliography{references}
\bibliographystyle{iclr2027_conference}

\clearpage
\appendix
\addtocontents{toc}{\protect\setcounter{tocdepth}{2}}
\renewcommand{\contentsname}{Appendix Contents}
\tableofcontents
\section{Data and model details}
This section gives what is needed to reproduce the data preparation and every component of \cref{sec:setup,sec:method}. \Cref{app:stats} defines the statistics and \cref{app:baselines} the decoders. The next three sections support the main results: what the brain adds to the camera path (\cref{app:neural}), the layout gain in every scene (\cref{app:corrected_scenes}) and the surfaces (\cref{app:surfaces}). \Cref{app:floors} gives the fMRI-free references and the range calibration, and \cref{app:corrected_pilot} the loss correction. \Cref{app:posthoc} holds post hoc robustness analyses, and \cref{app:posthoc_extra} the withheld scenes and low-level image features. \Cref{app:extended} lists additional analyses of the same data that are not included in this version, each with its outcome. Throughout, \ours{} includes the range calibration of \cref{sec:fusion}, and columns or rows marked $a=1$ are its uncalibrated ablation (\cref{app:range_ablation}).

\subsection{Data}
\label{app:data}

\paragraph{Stimuli.}
Each video is an 80-second first-person walkthrough of a furnished indoor scene at 30 frames per second.
The four styles of a scene (modern, Scandinavian, wabi-sabi and minimalist) share the floor plan and visit the rooms in the same order.
Their camera paths are not frame-synchronized, so pooling across styles only pairs videos whose poses agree at the same second (\cref{sec:pooling,app:pooling_diagnostics}).
Seven scenes have all four styles, and the others two or three (\cref{tab:app_rooms}).

\paragraph{Ground truth.}
For every second we cast rays from the recorded camera pose through a $384\times512$ pinhole grid with a 90\textdegree{} horizontal field of view and record the radial depth to the first surface of the 3D model, not the optical-axis depth.
Depth is stored at $96\times128$ and averaged in inverse depth to $48\times64$.
Depth Anything V2 predicts optical-axis depth $z$, converted to radial depth $r=z/u_z$ ($\vu$: unit camera ray) before downsampling.
Pixels closer than 5\,cm or without hits are invalid.

\paragraph{fMRI.}
Blood-oxygen-level-dependent (BOLD) responses are resampled to the 2\,mm MNI152NLin6Asym template and z-scored within each run.
The readout uses V1 (3{,}004 voxels), V2 (2{,}020), ventral V3 (V3v, 866) and human V4 (hV4, 1{,}160).
These are group regions from the Juelich maximum-probability atlas at a 25\% threshold, not retinotopic maps measured in each participant.
Targets are seconds 0 to 73, which leaves six seconds of video after the last target for the longest readout lag. Evaluation uses seconds 0 to 71, in nine non-overlapping eight-second routes.
Run time series are linearly interpolated at stimulus-locked one-second centers, matching the repetition time (TR) of 1\,s.

\paragraph{Cleaning.}
Before the stimulus windows are cut out, each run is detrended, z-scored and cleaned by regressing out 24 head-motion terms, cerebrospinal fluid and white-matter signals, five anatomical component-based noise correction (aCompCor) components and the available cosine terms.
These head-motion terms differ from the camera-motion regressors of the readout.
A run can contain several scenes, so this cleaning is not refit within each fold and can couple training and test responses.
No depth target enters it.
The results therefore describe offline decoding from run-cleaned responses, not independent preprocessing of a new scene.

\paragraph{Acquisition and participants.}
Data were acquired on a Siemens MAGNETOM Prisma at 3\,T with a HeadNeck\_20 coil (TR 1\,s, echo time 30\,ms, flip angle 55\textdegree{}, multiband factor 4, 60 slices of 2.5\,mm) and preprocessed with fMRIPrep 24.1.1 (Nipype 1.8.6, ANTs 2.5.3, FreeSurfer 7.3.2)~\citep{esteban2019fmriprep}.
Scenes 5 to 7 each have 74 of 76, and scene 8 has 38 of 39 possible recordings. The records do not say whether a missing recording was never acquired, interrupted or excluded.
Participant counts differ across scenes (6 to 19), so recording averages have unequal noise levels (matched-cohort controls in \cref{app:neural_controls}).

\begin{table}[ht]
  \centering
  \caption{\textbf{The ten scenes differ in styles and participants.} A recording is one participant watching one style video, and each video gives nine routes of eight consecutive one-second views. The participant total counts distinct people. Counts are those of \ours{}, and the decoders use two more recordings (scenes 9 and 10, \cref{app:baselines}).}
  \label{tab:app_rooms}
  \small
  \begin{tabular}{@{}lrrrrrrrrrr@{\hspace{1.2em}}r@{}}
    \toprule
    Scene & 1 & 2 & 3 & 4 & 5 & 6 & 7 & 8 & 9 & 10 & Total \\
    \midrule
    Archive ID suffix & 01 & 02 & 03 & 04 & 05 & 06 & 07 & 09 & 10 & 12 & \\
    \addlinespace[3pt]
    Styles & 4 & 4 & 4 & 4 & 4 & 4 & 4 & 3 & 2 & 3 & 36 \\
    Recordings & 24 & 24 & 24 & 24 & 74 & 74 & 74 & 38 & 24 & 36 & 416 \\
    Participants & 6 & 6 & 6 & 6 & 19 & 19 & 19 & 13 & 12 & 12 & 20 \\
    Routes & 36 & 36 & 36 & 36 & 36 & 36 & 36 & 27 & 18 & 27 & 324 \\
    \bottomrule
  \end{tabular}
\end{table}

\subsection{Model components}
\label{app:impl}

\paragraph{Layout code.}
The 22 entries are 12 wall distances, four floor extents, four opening ratios, and the median and interquartile range (IQR) of log depth.
Wall distances are the median log radial depth within 10\textdegree{} of the horizon in 12 bins of 7.5\textdegree{} over $[-45\text{\textdegree}, 45\text{\textdegree}]$.
Floor extents are the 95th percentile of log depth over pixels below $-15$\textdegree{} elevation in four vertical bands of equal width.
Opening ratios are the fraction of valid pixels in each quadrant that are more than twice as far as the median depth of the view.
Empty wall or floor bins fall back to the median, and empty openings to zero. Targets are z-scored with the mean and standard deviation of the nine training scenes of each fold.

\paragraph{Readout.}
The readout averages six ridge regressions,
\begin{equation}
  \hat\vy_t = \tfrac{1}{6}\textstyle\sum_{m=1}^{6}\big(\mW_m^\top[\mP_m \vx^{(m)}_t;\ \vk_t]+\vb_m\big).
  \label{eq:readout}
\end{equation}
Here $m$ indexes the members and $t$ the view. $\vx^{(m)}_t$ holds the 1{,}000 or 3{,}000 most reliable of the 7{,}050 voxels of early visual cortex (EVC). Their responses are averaged over 2 to 5 or 2 to 6 seconds after the view. $\mP_m$ projects them onto 16 principal components, and $\vk_t$ holds the 12 camera-motion regressors, also used at test time.
The six members use (voxels kept, ridge penalty, lag window) of (3000, 0.1, 2 to 5\,s), (1000, 0.3, 2 to 5\,s), (3000, 0.3, 2 to 5\,s), (1000, 0.1, 2 to 6\,s), (1000, 0.1, 2 to 5\,s) and (1000, 0.3, 2 to 6\,s).
Voxels are ranked by split-half reliability on the training scenes in the 3 to 6\,s window.
The ridge penalty is $\gamma N\mI$ on z-scored features, where $N$ is the number of training samples.
The readout with a \threegrid{} target (\cref{tab:main}) uses the same six ridges to predict the median log depth of each \threegrid{} block. It is a separate fit, not a rearrangement of the layout code.

\paragraph{Posterior spread.}
The posterior spread of an entry is the standard deviation of the readout's errors on a scene it was not trained on. To measure it without the held-out scene, we fit each readout on eight scenes, leaving out the held-out scene and one inner validation scene, and keep its errors on the validation scene. If the readout explains no training variance for an entry, its posterior mean and spread are the training mean and standard deviation.
With pooling across styles, spreads are estimated separately for each number of pooled videos, or overall below 2{,}000 samples.

\paragraph{Renderer.}
The renderer uses the eye height and pitch of the camera at each second.
Wall depth along a ray is $\exp(w(\theta))/\max(\cos\alpha,0.2)$, where $w(\theta)$ interpolates the wall code at azimuth $\theta$ and $\alpha$ is the elevation after pitch rotation. The code is measured near the horizon, so this only approximates a planar wall.
Floor depth is $h/(-d_y)$ for a ray with downward component $d_y<0$ and eye height $h$, capped by the floor extent of its band.
Openings blend each quadrant toward twice the median depth with a smooth step of half-width 6 pixels.
Depth is clipped to $[0.05, 30]$\,m.
The uncertainty map carries the posterior spreads $s_j$ of the code entries (\cref{sec:pooling}) through the renderer. Let $g$ be the dominant wall bin (or capped floor bin), with $s_g=0$ on the purely geometric floor. Let $q$ be the final inverse depth, $q_o$ the inverse depth after opening blending, $b$ the clipped global spread multiplier, and $z$ the log-depth deviation before correction divided by its IQR. Then
\begin{equation}
 (S^0)^2=q^2\left[(b s_g)^2+(z s_{21})^2+s_{20}^2\right]
 +\left(\frac{q\,s_o}{\max(q_o,10^{-6})}\right)^2,
 \quad s_o=|q_{\rm far}-q_{\rm base}|\sum_{k=1}^{4}M_k s_{16+k-1}.
\end{equation}
Indices of $s$ are zero-based, as in the code. $q_{\rm far}$ and $q_{\rm base}$ are the inverse depths of the opening target and of the unblended surface, and $M_k$ are smooth quadrant masks.
It omits covariance between entries, bin switches, quantiles and clipping, so the map is a reliability cue, not a calibrated posterior.

\paragraph{Generator.}
\label{app:generator_sampling}
The generator is a 3D U-Net with 5.9M parameters and 48, 96 and 144 channels at three resolutions. It downsamples space twice but never the eight views. It uses residual blocks with group normalization, and two attention blocks at the lowest resolution apply spatial and temporal self-attention (four heads).
Its input channels are the flow state $\vz_\tau$, the coarse depth $\mD^{0}$ and the uncertainty map $\mS^{0}$. The flow target $\vz_1$ is the residual between the true inverse depth and $\mD^0$, divided by one root-mean-square scalar $\rho$. The scalar $\rho$ is computed on the training scenes and frozen for test time.
The flow timestep and a 116-number vector condition the network through feature-wise linear modulation (FiLM), after a multilayer perceptron (MLP) of width 256. The vector holds the 22 posterior means and 22 posterior spreads, averaged over the eight views, and 72 motion features. They describe each of the eight views by its local position change, the sine and cosine of the heading change, pitch, eye-height change, speed and turning speed, and are standardized with fixed means and scales from 15{,}060 routes of ProcTHOR~\citep{deitke2022procthor}, a set of synthetic houses. This is the only external data. No weights pretrained on the Natural Scenes Dataset (NSD) or on external depth data are used.
We train with AdamW (learning rate $2\times10^{-4}$, weight decay $10^{-3}$), gradient clipping at 1.0, an exponential moving average of weights with decay 0.999, batch size 16 and bf16 precision for 20{,}000 steps. One fold takes about 15 minutes and 3.7\,GB of memory on one NVIDIA RTX 5090.
Training uses all 67 overlapping routes of each training video (2{,}144 to 2{,}278 routes per fold), and testing the nine non-overlapping routes of each video. We draw eight samples per recording with 48 Euler steps and average all samples of a video in inverse depth. The reciprocal of this average, in meters, is $\check{\mD}_t$.

\paragraph{Coordinate-consistent flow penalty.}
\label{app:flow_terms}
The penalty of \cref{eq:flow} compares the posterior with the code of the generated depth. The map $\mathcal A_i(q)=(\mathcal R^+_{\rm aff}(q)-\boldsymbol\eta_i)/\boldsymbol\sigma_i$ standardizes the fixed affine inverse of the renderer (\cref{app:loss_audit}). The inner-fold mean $\boldsymbol\eta_i$ and scale $\boldsymbol\sigma_i$ standardize the posterior mean of sampled recording $i$.
Weights are clipped precisions $w_j=\max(s_j,10^{-3})^{-2}$, normalized by the batch mean, and $\lambda=0.1$. The penalty is an approximation, not a calibrated likelihood.

\paragraph{Trajectory head and fusion head.}
For view $t$ with position $\vp_t$ and heading $\psi_t$, and for each of the 12 direction bins $b$ of the layout code, we compute the farthest distance the camera reaches,
\begin{equation}
  L_{t,b}=\max\big\{\|\vp_{t'}-\vp_t\| \,:\, |t'-t|\le K,\ \angle(\vp_{t'}-\vp_t)-\psi_t\in b\big\},
  \label{eq:tfs}
\end{equation}
which is zero for an empty bin. A distance reached by the camera is not an upper bound on the distance to a wall.
For each bin, the free-space features store $\log(1+L_{t,b})$, the number of path points in the bin and the time to the nearest one. They add the forward distance traveled over the previous and next $K$ seconds (the signed displacement along the current heading) and the signed turn over the next four seconds, for 39 values. With the 72 motion features, the trajectory head has 111 inputs.
Inner leave-one-scene-out validation chooses $K$ from $\{4, 8, 16, 37, 73\}$\,s and the ridge penalty from $\{1, 10, 10^2, 10^3, 10^4\}$. It maximizes layout $r$, with row order breaking ties. It selected $K=4$\,s in every scene we inspected, a penalty of $10^3$ for the trajectory head in every scene, and for the fusion head $10^3$ in eight scenes and $10^4$ in scenes 6 and 10.
The fusion head takes the 22 standardized pooled posterior means and the nine outputs of the trajectory head, standardized with training-scene statistics.
The low-pass filter of \cref{eq:lowfreq} is a separable Gaussian with $\sigma=4$ pixels and symmetric padding, applied to log depth.
The held-out scene is excluded throughout, but inner validation of the fusion head is not fully clean. It reuses frozen out-of-fold readout features, so the feature of one training scene can come from a readout trained on the inner validation scene. Removing this overlap would need a third level of folds that excludes all three scenes, which we did not fit.
The trajectory head on motion features alone (inputs M in \cref{tab:main}) is a frozen ridge on the 72 motion features. All layouts of \cref{tab:main} share the generator, which is conditioned on fMRI, so the surfaces of the trajectory head are not free of brain input.

\subsection{When pooling across styles applies}
\label{app:pooling_diagnostics}
Pooling helps only if partners exist and see similar depth, so we measure both. A video of another style is a partner at a second only if the two poses agree within 5\,cm in each position coordinate, 2\textdegree{} in heading and 0.5\textdegree/s in turning speed. Pitch is not constrained, and the largest pitch difference is 1.092\textdegree{}. When several videos qualify, the one with the most recordings is the partner, ties broken by video name, never by scores.
A partner is used at 1{,}473 of 2{,}592 video-seconds (56.8\%), from 0\% to 92.7\% by scene, and never in scene 9.
The depths of the two partners differ by a median of 0.314\,m (IQR 0.172 to 0.563\,m). This mixes small pose mismatch with changes in furniture.

\section{Evaluation and statistics}
\label{app:stats}
The ten held-out scenes are the units of the bootstrap and sign-flip tests. Response windows of 2 to 6\,s make neighboring views share BOLD samples, so per-view scores are not independent.
For a mean metric such as layout $r$ or pixel MAE, we take the difference between two methods on every route and average it within each scene.
We then resample the ten scene differences with replacement 20{,}000 times and report the ten-scene mean with the 2.5th and 97.5th percentiles (paired scene bootstrap).
For rate metrics such as row order, each scene pools its eligible cell pairs before differencing.
Main-table layout and pixel-MAE point estimates instead pool all routes or pairs (route-pooled values), so they can differ slightly from ten-scene means.
All intervals are descriptive 95\% intervals without correction for multiple comparisons. All ten scenes were used during development, and each model was trained with one seed. Pose noise was tested at registration (\cref{app:registration_drift}) and, post hoc, in the camera-path inputs (\cref{app:posthoc}). Single recordings with the corrected generator were tested only post hoc (\cref{app:posthoc}). Filmed real scenes, single-participant models and unobserved surfaces were not tested.
Each analysis uses one fixed bootstrap seed, released with the code.
Intervals of the historical models, trained before the loss correction (\cref{app:loss_audit}), were route-weighted and are not combined with the corrected contrasts.

\paragraph{Sign-flip tests.}\label{app:signflip}
Percentile intervals from ten scenes can be too narrow. For four main layout-$r$ contrasts we therefore also run an exact two-sided sign-flip test. The ten scene differences are multiplied by all $2^{10}$ sign patterns, and $p$ is the fraction of patterns whose absolute mean is at least the observed one. With ten scenes, the smallest possible value is 0.002.
The contrasts are the \ours{} layout minus the generator alone ($p=0.002$), minus the trajectory head ($p=0.002$) and minus the fixed template ($p=0.004$), and Brain-Diffuser + \ours{} layout minus Brain-Diffuser + trajectory layout ($p=0.002$). After Holm correction for these four tests, the largest value is 0.008.
The same test gives $p=0.002$ for the four brain-input controls of \cref{sec:complementary}: feature replacement in layout $r$ and in F-score at $a=2$, the circular shift in F-score at $a=2$, and EVC minus the camera-motion regressors alone in cosine similarity. Each favors the intact input in all ten scenes.
All these contrasts were named after the results were known and were not preregistered.

\subsection{Metric definitions}
\label{app:metric_definitions}
Row order is the fraction of cell pairs in the same row of the \threegrid{} layout that the prediction orders correctly. Only pairs whose true log depths differ by at least 0.2 (about 22\% in depth) count. Comparing cells within a row separates left-to-right structure from the near-floor and far-wall contrast that all indoor views share. Left/right uses the same threshold.
Layout $r$ is averaged over views and then over routes, so it measures agreement within views, not a pooled correlation.
The surface code voxel-merges each route separately, but four routes at a time share a grid origin one meter below their joint minimum on each coordinate. The origin depends on the batch, so reproductions must keep the saved route order and batch size.
Let $P$ be the voxel-merged predicted points and $Q$ the valid ground-truth ray hits from the same eight poses, not voxel-merged. With $d(p,Q)=\min_{q\in Q}\|p-q\|_2$,
\begin{align}
 C(P,Q)&=\tfrac12\left[|P|^{-1}\sum_{p\in P}d(p,Q)+|Q|^{-1}\sum_{q\in Q}d(q,P)\right],\\
 p_\delta&=|P|^{-1}\sum_{p\in P}\mathbf1[d(p,Q)\le\delta],\qquad
 r_\delta=|Q|^{-1}\sum_{q\in Q}\mathbf1[d(q,P)\le\delta],\\
 F_\delta&=\frac{2p_\delta r_\delta}{p_\delta+r_\delta},\qquad\delta=0.25\,\mathrm m.
\end{align}
The Chamfer distance $C$ uses unsquared distances in meters. $p_\delta$ and $r_\delta$ are precision and recall, and $F_\delta$ is the F-score.
Pixel MAE compares predicted radial depth with ground truth on valid pixels.

\section{Brain-decoding baselines}
\label{app:baselines}
All decoders share the folds, routes, targets and metrics of \ours{}, and every trainable part is fit on the nine training scenes of each fold.

\paragraph{Protocol.}
Decoders read participant-averaged responses, averaged over 2 to 6 seconds after each view, from seven regions of interest (ROIs). These are V1, V2, V3v and hV4 in early visual cortex and the scene-selective occipital place area (OPA), parahippocampal place area (PPA) and retrosplenial complex (RSC). A method that prescribes early visual areas uses V1 to hV4, and one that prescribes higher visual areas uses hV4, PPA, OPA and RSC.
Each view's video frame is resized to the input size of each model without cropping.
Reconstructed images are converted to metric depth with Depth Anything V2 (DA V2) Metric Indoor Large~\citep{yang2024depthv2}. It overestimates depth on our rendered scenes by about a factor of 1.75, so we correct its scale with a log-affine map fitted on the frames of the training scenes (DA V2 scale calibration).
Predictions were frozen before scoring under a written plan.
The decoders read responses averaged over 418 recordings, and \ours{} over 416. The two extra recordings (one participant, minimalist style of scenes 9 and 10) have no second style to pool with. The trajectory layout uses no fMRI, and decoders with the \ours{} layout combine the two sets of responses.

\paragraph{Methods.}
\emph{Linear depth decoding} maps fMRI to the first 64 principal components of log depth with ridge regression~\citep{hoerl1970ridge}, a standard linear approach to fMRI decoding~\citep{naselaris2011encoding}.
\emph{CLIP retrieval} maps fMRI to the image embedding of CLIP (Contrastive Language-Image Pre-training) ViT-L/14~\citep{radford2021learning} with ridge regression, a simple analogue of the contrastive retrieval branch of MindEye~\citep{scotti2023reconstructing}, and averages the depth maps of the ten nearest training frames.
\emph{Image decoding} follows \citet{takagi2023high}. Early visual cortex predicts the Stable Diffusion latent, and higher visual cortex predicts the text embedding of a caption. \citet{takagi2023high} averaged the five human-written captions of each image in the COCO dataset. Video frames have no captions, so we instead caption each frame with BLIP, an image captioning model~\citep{li2022blip}. Image-to-image diffusion then generates the image (strength 0.8, 50 denoising diffusion implicit model steps, guidance 5.0). \emph{Depth decoding}~\citep{takagi2023improving} predicts hidden states of the dense prediction transformer DPT-large~\citep{ranftl2021vision}, whose head outputs relative depth calibrated on the training scenes.
\emph{Brain-Diffuser}~\citep{ozcelik2023natural} predicts latents of a very deep variational autoencoder (VDVAE)~\citep{child2021very} and CLIP image and text embeddings that guide Versatile Diffusion~\citep{xu2023versatile} (strength 0.75, mixing 0.4). It keeps its released step that re-scales its predictions to their mean and spread over the held-out videos. This step uses no labels, but it uses all held-out predictions at once (it is transductive).
\emph{MindEye}~\citep{scotti2023reconstructing} trains an MLP (width 1024, 64 epochs per fold) to predict CLIP ViT-L/14 hidden states, and a low-level branch starts Versatile Diffusion image variation (strength 0.85). Our version omits the ConvNeXt contrastive term and uses the Versatile Diffusion autoencoder for low-level targets.
We did not reproduce \citet{gaziv2021more}, full DREAM~\citep{xia2024dream} or MinD-Video. Our adaptations decode each view alone, so they cannot show that \ours{} beats temporal fMRI-to-video decoders.

\paragraph{Regularization.}
On the penalty grids of \citet{takagi2023high} and of depth decoding, every fold chose the largest penalty, so the best one could lie outside these grids. With these penalties image decoding also produced textureless color fields. We therefore report the stronger ``wider ridge'' variants, fit on a wider grid. For linear depth, a wider grid selected the same penalty.

\paragraph{Image similarity.}
We test whether CLIP ViT-L/14 embeddings pick the viewed frame over another frame of the same held-out scene. This two-way identification is 0.500 and 0.525 for the official and wider-ridge versions of \citet{takagi2023high}, 0.627 for Brain-Diffuser and 0.624 for MindEye (chance 0.5).
The grayscale structural similarity (SSIM) of the wider-ridge decoded images to the viewed frame is 0.651, against 0.649 for another frame of the same scene.
So the images are indoor scenes that match their specific view weakly at best (\cref{app:extended}).

\subsection{DREAM depth branch}
\label{app:dream_adapter}
We adapt the released depth-regression branch of DREAM~\citep{xia2024dream}, which uses a VDVAE pretrained on ImageNet.
In each fold, ridge regression (penalty fixed at 50{,}000) maps the fMRI of the training scenes to VDVAE latents of depth images, coded as gray level $\log(d/0.05)/\log(600)$ at $64\times64$. Eight decoded samples are averaged per view.
Two differences remain. The DREAM depth branch has no camera input, and it reads the decoders' responses, which average two more recordings.
The primary contrast, fixed in advance, is the DREAM depth branch minus the generator alone in Chamfer distance at $a=1$. It is $+0.134$\,m [0.097, 0.175], and all ten scenes favor the generator. F-score is also lower in all ten scenes.
The DREAM depth branch has the better layout $r$ ($+0.128$ [0.087, 0.158], nine scenes) and is better on most other depth-map layout measures (\cref{app:extended}). The differences in input and cohort prevent attributing the gap to the architecture.

\paragraph{Range ratio.}
The range ratio measures how much of the true depth spread a prediction keeps (\cref{sec:surface}). It divides the standard deviation of the predicted $3{\times}4$ cell log depths by that of the true ones, so its ideal value is 1. Before range calibration, the route-pooled range ratios of our ridge layouts are 0.46 for the readout with a \threegrid{} target, 0.50 for the trajectory head from motion, 0.54 for the trajectory head with free space and 0.49 for \ours{}. The ratio is 0.78 for the generator, and DA V2 gives Brain-Diffuser and MindEye 0.84. Range calibration raises the ratio of \ours{} to 0.97. \Cref{app:extended} summarizes ten-scene means instead.

\subsection{Post hoc layout transfer to all seven decoders}
\label{app:transfer}
After the two planned combinations (\cref{app:layout_guidance}) were scored, we asked whether the layout guides every decoder and applied the same low-frequency replacement (\cref{eq:lowfreq}) to the frozen depth of all seven decoders, with the trajectory layout or the \ours{} layout, and scored each version at $a=1$ and $a=2$ (every score and paired difference omitted; \cref{app:extended}). The transfer's Brain-Diffuser and linear-depth rows reproduce the planned combinations exactly, and its $a=2$ versions form the ``+ path'' and ``+ \ours{}'' columns of \cref{tab:baselines}.

With the \ours{} layout, every decoder reaches a layout $r$ of 0.487 to 0.493, whatever its own value (0.171 to 0.429). Over the trajectory layout, the \ours{} layout raises layout $r$ by 0.036 to 0.041 and lowers pixel MAE by 0.077 to 0.078\,m, in all ten scenes for each decoder (sign-flip $p=0.002$ each). Row order and left/right gain less (0.007 to 0.015), and no sign-flip $p$ is below 0.05.

The surfaces are mixed. At $a=1$, the replacement lowers the F-score of the three decoders whose surfaces are best alone (Brain-Diffuser, MindEye and CLIP retrieval) in at least eight of ten scenes each, and raises their Chamfer distance in at least seven. It helps the two with the weakest surfaces, depth decoding and the DREAM depth branch. At $a=2$, every decoder except linear depth has a higher F-score with the \ours{} layout than alone. But the factor was fixed for the layout rows, and it hurts the unmodified decoders. Unmodified Brain-Diffuser, for example, has an F-score of 0.492 at $a=1$ and 0.424 at $a=2$. So these contrasts do not show that the layout beats each decoder at its own best factor. For Brain-Diffuser, that comparison is within noise (\cref{app:layout_guidance}).
The brain part of the \ours{} layout adds little to the surfaces. Over the trajectory layout, the F-score changes by $+0.017$ to $+0.020$ at $a=1$, with every interval including zero, and by $+0.014$ to $+0.023$ at $a=2$. The Chamfer distance at $a=2$ falls by 0.018 to 0.030\,m in seven to nine scenes, but the sign-flip $p$ is above 0.05 for linear depth, depth decoding and the DREAM depth branch. The exact sign-flip test is the conservative reading.

Panel (a) of \cref{tab:baselines_ci} compares \ours{} itself with each decoder under the three conditions of \cref{tab:baselines}. Given the same camera path, that is, with the trajectory layout, every decoder has a lower layout $r$ than \ours{}, in eight to ten scenes (sign-flip $p\le0.010$). Linear depth, image decoding, depth decoding and the DREAM depth branch also have a lower F-score (eight to ten scenes, $p\le0.016$). The F-scores of CLIP retrieval, Brain-Diffuser and MindEye are slightly lower but within noise (six scenes each, $p\ge0.22$). We added these sign-flip tests after the results were known.

Panel (b) of \cref{tab:baselines_ci} gives the gains behind the markers of \cref{tab:baselines}. Over its unmodified output, every decoder has a positive mean gain with the \ours{} layout in all three metrics. The intervals exclude zero in 17 of 21 cases (sign-flip $p\le0.037$). The four exceptions are the F-score and Chamfer distance of Brain-Diffuser and MindEye, whose gains are small ($p\ge0.35$). These gains include range calibration, which the unmodified output lacks. When both are calibrated, the \ours{} layout does not raise the F-score of linear depth over the decoder alone ($-0.002$ at $a=2$, \cref{app:extended}). \Cref{tab:baselines} omits pixel MAE. \Cref{tab:main} gives it for \ours{}. It is 0.501\,m for depth from the video frames and 1.188\,m for the fixed template.

\begin{table}[ht]
  \centering
  \caption{\textbf{Given the same camera path, \ours{} has a higher layout $r$ than every decoder, and the \ours{} layout improves every decoder.} (a) \ours{} minus each method of \cref{tab:baselines}. (b) The markers of \cref{tab:baselines}: decoder with the \ours{} layout and range calibration ($a=2$) minus unmodified decoder ($a=1$), $^\ast$ if the interval excludes zero. The surface gains of Brain-Diffuser and MindEye are not reliable. Ten-scene means of per-scene differences, 95\% paired scene-bootstrap intervals and scenes of ten favoring \ours{} or the layout. Post hoc exact sign-flip $p$ (\cref{app:signflip}) for (b) and the trajectory-layout rows of (a). Layout rows are post hoc except for linear depth and Brain-Diffuser (\cref{app:transfer}). $^\ddagger$DA V2 on the frames. $^\dagger$Our adaptation (\cref{app:dream_adapter}).}
  \label{tab:baselines_ci}
  \footnotesize
  \setlength{\tabcolsep}{4pt}
  \newcommand{\cit}[2]{{\scriptsize[#1,\,#2]}}
  \begin{tabular}{@{}l r@{\,}l@{\,}r c r@{\,}l@{\,}r c@{}}
    \multicolumn{9}{@{}l}{\textit{(a) \ours{} minus each method}} \\
    \toprule
    & \multicolumn{4}{c}{Layout $r$\,\up} & \multicolumn{4}{c}{F@25cm\,\up} \\
    \cmidrule(lr){2-5}\cmidrule(l){6-9}
    \ours{} minus & \multicolumn{3}{c}{Mean [95\% CI] (scenes)} & $p$ & \multicolumn{3}{c}{Mean [95\% CI] (scenes)} & $p$ \\
    \midrule
    \multicolumn{9}{@{}l}{\emph{References}} \\
    Video frames + DA V2$^\ddagger$ & $-$0.393 & \cit{$-$0.445}{$-$0.340} &  & -- & $-$0.258 & \cit{$-$0.295}{$-$0.221} &  & -- \\
    Fixed template & $+$0.144 & \cit{0.084}{0.204} &  & -- & $+$0.093 & \cit{0.080}{0.108} &  & -- \\
    \midrule
    \multicolumn{9}{@{}l}{\emph{Decoder alone (unmodified output)}} \\
    Linear depth & $+$0.045 & \cit{0.011}{0.079} & (7) & -- & $+$0.091 & \cit{0.066}{0.113} & (10) & -- \\
    CLIP retrieval & $+$0.210 & \cit{0.157}{0.267} & (10) & -- & $+$0.029 & \cit{0.014}{0.044} & (8) & -- \\
    Image decoding & $+$0.300 & \cit{0.259}{0.335} & (10) & -- & $+$0.042 & \cit{0.021}{0.062} & (9) & -- \\
    Depth decoding & $+$0.156 & \cit{0.123}{0.193} & (10) & -- & $+$0.167 & \cit{0.136}{0.194} & (10) & -- \\
    DREAM depth branch$^\dagger$ & $+$0.045 & \cit{0.012}{0.075} & (7) & -- & $+$0.116 & \cit{0.086}{0.143} & (10) & -- \\
    Brain-Diffuser & $+$0.152 & \cit{0.128}{0.175} & (10) & -- & $-$0.008 & \cit{$-$0.027}{0.011} & (4) & -- \\
    MindEye & $+$0.206 & \cit{0.168}{0.249} & (10) & -- & $-$0.010 & \cit{$-$0.023}{0.004} & (4) & -- \\
    \midrule
    \multicolumn{9}{@{}l}{\emph{+ trajectory layout (``+ path''), the same camera path as \ours{}}} \\
    Linear depth & $+$0.028 & \cit{0.016}{0.039} & (9) & 0.006 & $+$0.044 & \cit{0.028}{0.060} & (10) & 0.002 \\
    CLIP retrieval & $+$0.031 & \cit{0.017}{0.044} & (8) & 0.008 & $+$0.011 & \cit{$-$0.005}{0.026} & (6) & 0.223 \\
    Image decoding & $+$0.032 & \cit{0.020}{0.044} & (10) & 0.002 & $+$0.028 & \cit{0.012}{0.043} & (8) & 0.016 \\
    Depth decoding & $+$0.027 & \cit{0.015}{0.038} & (9) & 0.006 & $+$0.049 & \cit{0.032}{0.065} & (10) & 0.002 \\
    DREAM depth branch$^\dagger$ & $+$0.032 & \cit{0.020}{0.044} & (9) & 0.004 & $+$0.049 & \cit{0.032}{0.065} & (10) & 0.002 \\
    Brain-Diffuser & $+$0.024 & \cit{0.011}{0.037} & (8) & 0.010 & $+$0.008 & \cit{$-$0.007}{0.021} & (6) & 0.348 \\
    MindEye & $+$0.031 & \cit{0.018}{0.045} & (9) & 0.006 & $+$0.008 & \cit{$-$0.006}{0.023} & (6) & 0.328 \\
    \midrule
    \multicolumn{9}{@{}l}{\emph{+ \ours{} layout}} \\
    Linear depth & $-$0.009 & \cit{$-$0.013}{$-$0.006} & (1) & -- & $+$0.029 & \cit{0.024}{0.033} & (10) & -- \\
    CLIP retrieval & $-$0.008 & \cit{$-$0.014}{$-$0.002} & (2) & -- & $-$0.009 & \cit{$-$0.014}{$-$0.004} & (2) & -- \\
    Image decoding & $-$0.008 & \cit{$-$0.013}{$-$0.003} & (1) & -- & $+$0.010 & \cit{0.006}{0.014} & (9) & -- \\
    Depth decoding & $-$0.011 & \cit{$-$0.016}{$-$0.006} & (1) & -- & $+$0.035 & \cit{0.029}{0.041} & (10) & -- \\
    DREAM depth branch$^\dagger$ & $-$0.008 & \cit{$-$0.012}{$-$0.004} & (1) & -- & $+$0.035 & \cit{0.030}{0.040} & (10) & -- \\
    Brain-Diffuser & $-$0.012 & \cit{$-$0.018}{$-$0.007} & (1) & -- & $-$0.015 & \cit{$-$0.022}{$-$0.007} & (1) & -- \\
    MindEye & $-$0.007 & \cit{$-$0.013}{$-$0.001} & (2) & -- & $-$0.015 & \cit{$-$0.022}{$-$0.008} & (1) & -- \\
    \bottomrule
  \end{tabular}

  \vspace{8pt}
  \scriptsize
  \setlength{\tabcolsep}{1.9pt}%
  \def\sig{\textsuperscript{\fontsize{6}{6}\selectfont$\ast$}}\def\nsig{\phantom{\sig}}%
  \begin{tabular}{@{}l r@{\,}l@{\;}r c@{\hspace{7pt}}r@{\,}l@{\;}r c@{\hspace{7pt}}r@{\,}l@{\;}r c@{}}
    \multicolumn{13}{@{}l}{\textit{(b) Decoder + \ours{} layout ($a=2$) minus the decoder alone ($a=1$), the markers of \cref{tab:baselines}}} \\
    \toprule
    & \multicolumn{4}{c}{Layout $r$\,\up} & \multicolumn{4}{c}{F@25cm\,\up} & \multicolumn{4}{c}{Chamfer (m)\,\down} \\
    \cmidrule(lr){2-5}\cmidrule(lr){6-9}\cmidrule(l){10-13}
    Decoder & \multicolumn{3}{c}{Gain [95\% CI] (scenes)} & $p$ & \multicolumn{3}{c}{Gain [95\% CI] (scenes)} & $p$ & \multicolumn{3}{c}{Gain [95\% CI] (scenes)} & $p$ \\
    \midrule
    Linear depth & $+$0.054\sig & \cit{0.021}{0.088} & (7) & 0.020 & $+$0.062\sig & \cit{0.041}{0.081} & (9) & 0.004 & $+$0.072\sig & \cit{0.029}{0.109} & (9) & 0.010 \\
    CLIP retrieval & $+$0.219\sig & \cit{0.166}{0.274} & (10) & 0.002 & $+$0.038\sig & \cit{0.023}{0.053} & (10) & 0.002 & $+$0.038\sig & \cit{0.012}{0.059} & (9) & 0.020 \\
    Image decoding & $+$0.308\sig & \cit{0.268}{0.343} & (10) & 0.002 & $+$0.031\sig & \cit{0.009}{0.053} & (7) & 0.037 & $+$0.042\sig & \cit{0.017}{0.067} & (7) & 0.016 \\
    Depth decoding & $+$0.166\sig & \cit{0.134}{0.204} & (10) & 0.002 & $+$0.132\sig & \cit{0.104}{0.155} & (10) & 0.002 & $+$0.283\sig & \cit{0.219}{0.344} & (10) & 0.002 \\
    DREAM depth branch$^\dagger$ & $+$0.053\sig & \cit{0.021}{0.083} & (7) & 0.025 & $+$0.081\sig & \cit{0.054}{0.105} & (9) & 0.004 & $+$0.115\sig & \cit{0.065}{0.162} & (9) & 0.006 \\
    Brain-Diffuser & $+$0.164\sig & \cit{0.140}{0.188} & (10) & 0.002 & $+$0.006\nsig & \cit{$-$0.009}{0.022} & (5) & 0.457 & $+$0.011\nsig & \cit{$-$0.012}{0.032} & (7) & 0.354 \\
    MindEye & $+$0.213\sig & \cit{0.171}{0.259} & (10) & 0.002 & $+$0.005\nsig & \cit{$-$0.007}{0.017} & (5) & 0.486 & $+$0.005\nsig & \cit{$-$0.020}{0.025} & (8) & 0.713 \\
    \bottomrule
  \end{tabular}
\end{table}

\section{Brain-input controls and cortical regions}
\label{app:neural}
These controls ask what the brain adds to camera motion, which cortical input carries layout information and whether the pipeline depends on it. None tests biological causation.

\subsection{Camera motion, free space and EVC}
One test of what EVC adds to camera motion keeps the motion input fixed. We fit the readout with a \threegrid{} target, on the same folds, to the 12 camera-motion regressors alone (motion alone). It reaches a layout $r$ of 0.338 (route-pooled), and adding EVC raises it by 0.098 [0.053, 0.151], in nine scenes ($p=0.004$).
A second comparison starts from the trajectory head on the 72 motion features alone (layout $r$ 0.390, \cref{tab:main}). Two alternatives beat it by the same amount, each in eight scenes: the readout with a \threegrid{} target from EVC and the 12 regressors, by 0.050 [0.015, 0.092] ($p=0.025$), and the trajectory head with free space added to the motion features, by 0.050 [0.022, 0.076] ($p=0.012$). The readout also lowers pixel MAE by 0.117\,m [0.061, 0.172] in eight scenes ($p=0.010$). The two read different motion inputs, so this comparison alone does not isolate EVC.

\subsection{Matched-budget comparison of cortical regions}
\label{app:roi_readout}
To test the choice of EVC (\cref{sec:readout}), we compare EVC (V1, V2, V3v and hV4), scene-selective regions (OPA, PPA and RSC) and their union under one readout fixed before scoring: 1{,}000 most reliable voxels, 16 principal components, ridge penalty $\gamma=0.1$ and a 2 to 5\,s lag window. Voxel ranking, principal components and standardization use only the nine training scenes of each fold. Every readout includes the 12 camera-motion regressors, and a ridge on these alone is the reference. The comparison tests the readout alone, without a noise ceiling.

EVC predicts the code best (\cref{tab:roi_readout}). In cosine similarity, it beats the scene-selective regions by 0.160, in nine scenes, and these still beat motion alone by 0.059, also in nine scenes (intervals in \cref{tab:contrasts,tab:roi_readout}). Adding the scene-selective regions to EVC changes little: the union minus EVC is $-0.0003$ [$-0.0030$, 0.0018] in cosine similarity, and the union's mean squared error is lower by only 0.0010 [0.0002, 0.0021]. This is expected, because the union's reliability ranking keeps 979 to 992 EVC voxels and only 8 to 21 scene-selective voxels per fold (PPA 5 to 18, OPA 0 to 3, RSC 0 to 2). The union therefore does not test a separate budget for the scene-selective regions, and no comparison shows that a region is necessary. For every input, wall distances are the best predicted of the four groups of entries, a property of these targets and this readout.

The responses used here average 25 and 37 recordings in scenes 9 and 10, against 24 and 36 for \ours{}, so this is not a full-pipeline ablation with one cohort. Camera-motion regressors do not control image motion from texture, occlusion and parallax, so it shows predictive information, not a 3D representation independent of image features~\citep{shafer2026gabor}.

\begin{figure}[ht]
  \centering
  \includegraphics[width=\linewidth]{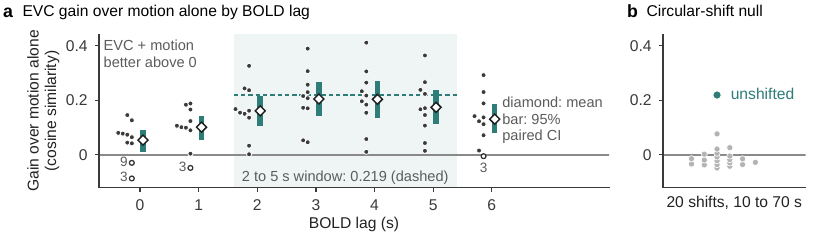}
  \caption{\textbf{EVC's gain over camera motion peaks at a 3 to 4\,s lag and exceeds all 20 circular shifts.} Gain: cosine similarity of the EVC readout minus that of motion alone. \textbf{(a)} Single-second BOLD lags. Dots: scenes (hollow and numbered at or below zero). Diamonds and bars: mean and 95\% paired scene-bootstrap interval. Shading and dashed line: the 2 to 5\,s window of the main readout, which averages four lags and can gain more than any single one. \textbf{(b)} Circular shifts of 10 to 70\,s (gray) and the unshifted window (teal), on the scale of (a).}
  \label{fig:readout_lag}
\end{figure}

\paragraph{Lag profile and circular-shift control.} A gain carried by the BOLD response should peak a few seconds after the stimulus and vanish when the BOLD series is shifted. We refit the readout with single-second lags from 0 to 6\,s and, for the 2 to 5\,s window, after circular shifts of the BOLD series by 20 offsets from 10 to 70\,s (\cref{fig:readout_lag}). The gain over motion alone is largest at 3 and 4\,s, in all ten scenes (0.205 [0.143, 0.266] and 0.204 [0.135, 0.271]). It is smallest at 0\,s, at 0.054 [0.010, 0.091], in eight scenes. Some gain at 0\,s is expected, because the layout code changes slowly and the hemodynamic response is spread in time. The unshifted gain (0.219) exceeds all 20 shifted gains (largest 0.077), a permutation $p$ of 1/21. The pooling partner does not enter this readout.

\begin{table}[ht]
 \centering\small
 \caption{\textbf{With 1{,}000 voxels each, EVC predicts the layout code best, and the scene-selective regions beat motion alone.} Ten-scene means. Cos.: cosine similarity of centered predicted and true codes. Entry $r$: mean per-entry correlation (neither is layout $r$). MSE, MAE: mean squared and absolute error (std.: standardized units). $\Delta$Cos.: minus motion alone, with 95\% paired scene-bootstrap interval. Scenes $>0$: of 10, not applicable to the reference row. Bold: within 0.002 of best. Contrasts between regions: \cref{app:roi_readout} and \cref{tab:contrasts}.}
 \label{tab:roi_readout}
 \begin{tabular}{@{}lcccccc@{}}
 \toprule
 Input & Cos.\,\up & Entry $r$\,\up & \begin{tabular}[b]{@{}c@{}}MSE\,\down\\(std.)\end{tabular} & \begin{tabular}[b]{@{}c@{}}MAE\,\down\\(std.)\end{tabular} & $\Delta$Cos.\,\up & \begin{tabular}[b]{@{}c@{}}Scenes\\$>0$\end{tabular} \\
 \midrule
 EVC (V1 to hV4) & \best{0.334} & \best{0.394} & \best{0.848} & \best{0.678} & $+$0.219 \ci{$+$0.150}{$+$0.290} & 10 \\
 Scene-selective & 0.174 & 0.221 & 0.971 & 0.740 & $+$0.059 \ci{$+$0.019}{$+$0.088} & 9 \\
 Union (both) & \best{0.334} & \best{0.394} & \best{0.847} & \best{0.678} & $+$0.219 \ci{$+$0.151}{$+$0.287} & 10 \\
 \midrule
 Motion alone (reference) & 0.115 & 0.185 & 0.987 & 0.751 & -- & -- \\
 \bottomrule
 \end{tabular}
\end{table}

\begin{table}[ht]
\centering\fontsize{8}{10}\selectfont
\caption{\textbf{Replacing or shifting the brain input worsens layout and surfaces in most scenes.} Intact minus control, for feature replacement (readout kept) and circular shift (readout refitted). Ten-scene means with 95\% paired scene-bootstrap intervals (low, high). Scenes better: of ten, scenes in which the intact input scores better (per-scene values omitted; \cref{app:extended}). Pixel MAE and Chamfer in cm (m in the other tables), F in percentage points (pp). G: generator, D: \ours{}, subscript: $a$. Only one mean favors the control, G$_2$ Chamfer under feature replacement, within noise.}
\label{tab:neural_controls_per_scene}
\setlength{\tabcolsep}{2.3pt}
\begin{tabular}{@{}l rr@{\hspace{10pt}}rrrr@{\hspace{10pt}}rrrr@{\hspace{10pt}}rrrr@{}}
\toprule
 & \multicolumn{2}{c}{Layout $r$ \up} & \multicolumn{4}{c}{Pixel MAE (cm) \down} & \multicolumn{4}{c}{F@25cm (pp) \up} & \multicolumn{4}{c}{Chamfer (cm) \down}\\
\cmidrule(lr){2-3}\cmidrule(lr){4-7}\cmidrule(lr){8-11}\cmidrule(lr){12-15}
 & G & D & G$_1$ & G$_2$ & D$_1$ & D$_2$ & G$_1$ & G$_2$ & D$_1$ & D$_2$ & G$_1$ & G$_2$ & D$_1$ & D$_2$\\
\midrule
\multicolumn{15}{@{}l}{\emph{Feature replacement: intact minus centered EVC features}}\\
Mean & $+$0.064 & $+$0.037 & $-$5.7 & $-$3.5 & $-$7.7 & $-$4.2 & $+$2.1 & $+$1.1 & $+$4.0 & $+$3.6 & $-$2.3 & $+$0.3 & $-$4.7 & $-$2.5\\
\quad 95\% low & $+$0.046 & $+$0.025 & $-$7.4 & $-$5.8 & $-$10.6 & $-$6.6 & $+$1.4 & $+$0.7 & $+$2.2 & $+$2.4 & $-$3.0 & $-$0.4 & $-$6.4 & $-$3.6\\
\quad 95\% high & $+$0.082 & $+$0.051 & $-$4.4 & $-$1.4 & $-$5.3 & $-$2.0 & $+$2.8 & $+$1.5 & $+$5.9 & $+$5.0 & $-$1.7 & $+$0.9 & $-$3.1 & $-$1.1\\
Scenes better & 10 & 10 & 10 & 7 & 10 & 9 & 10 & 10 & 10 & 10 & 10 & 4 & 10 & 9\\
\midrule
\multicolumn{15}{@{}l}{\emph{Circular shift: intact minus shifted primary BOLD}}\\
Mean & $+$0.078 & $+$0.043 & $-$6.9 & $-$6.9 & $-$8.4 & $-$6.8 & $+$2.5 & $+$1.6 & $+$4.2 & $+$4.4 & $-$2.8 & $-$0.8 & $-$5.2 & $-$3.6\\
\quad 95\% low & $+$0.047 & $+$0.028 & $-$9.1 & $-$9.5 & $-$11.8 & $-$10.6 & $+$1.6 & $+$1.1 & $+$2.5 & $+$2.9 & $-$3.8 & $-$1.4 & $-$7.0 & $-$5.3\\
\quad 95\% high & $+$0.112 & $+$0.059 & $-$5.0 & $-$4.4 & $-$5.6 & $-$4.2 & $+$3.3 & $+$2.2 & $+$6.1 & $+$6.0 & $-$2.0 & $-$0.1 & $-$3.5 & $-$2.0\\
Scenes better & 10 & 10 & 10 & 10 & 10 & 10 & 10 & 10 & 10 & 10 & 10 & 8 & 10 & 9\\
\bottomrule
\end{tabular}
\end{table}

\subsection{Replacing the EVC features}
\label{app:feature_zero}
To test whether the fitted pipeline uses the EVC features it receives at test time, for each recording and its pooling partner we set the scores of the EVC principal components to their training-scene center and keep the fitted readout. Camera-motion regressors, model weights, pooling partners, posterior spreads, trajectory layouts and reliability masks stay fixed. We recompute the posterior means and everything that depends on them: coarse depth, uncertainty maps and \ours{} layouts. With intact (unaltered) features, the layout $r$ of \ours{} is higher by 0.0373 and its F-score at $a=2$ by 0.0363, in all ten scenes. Its pixel MAE falls by 0.0418\,m and its Chamfer distance by 0.0251\,m, both in nine scenes (intervals in \cref{tab:neural_controls_per_scene}). In scene 1, pixel MAE and Chamfer distance at $a=2$ are worse, although layout $r$ and F-score improve. For the generator alone, the Chamfer change at $a=2$ is +0.0028\,m [$-0.0037$, 0.0091], within noise. This control is not a shuffle of the raw BOLD signal. The posterior spreads keep information from the intact fit, and centered features can combine unusually with the unchanged camera-motion regressors.

\subsection{Circular shift of the primary BOLD signal}
\label{app:circular_corrected}
Feature replacement keeps the fitted readout. To test whether the full pipeline also depends on the timing of the BOLD signal, we circularly shift the primary BOLD signal (that of the evaluated recording) by 37 of 80 positions before lag averaging and fit a separate principal component analysis and ridge readout on the shifted data. The camera-motion regressors are not shifted. Voxel rankings, posterior spreads and masks from the intact data are kept, and we regenerate every dependent prediction on 324 matched eight-view routes. The primary contrast, fixed in advance, is the F-score of \ours{} at $a=2$. The intact signal raises it from 0.4403 to 0.4840 (by 0.0436) in all ten scenes (interval in \cref{tab:contrasts}). Chamfer distance favors the intact signal on average but not in every scene. At $a=2$, intact BOLD gives the generator a worse Chamfer distance in scenes 2 and 9, and \ours{} a worse one in scene 1. At $a=1$, both improve in all scenes. Layout $r$ does not depend on $a$, so its two contrasts are one result. Because the readout is refitted and information from the intact data remains, this is not a complete null control. \Cref{tab:neural_controls_per_scene} summarizes both controls.

\subsection{Historical controls: single recordings and matched cohorts}
\label{app:neural_controls}
Two earlier controls agree with the corrected ones: the intact signal still helps when fewer recordings are averaged at inference. Both used the historical generator (before the loss correction, \cref{app:loss_audit}), no pooling across styles and the same 37-position shift. With one recording per reconstruction (readout still trained on averaged data), intact BOLD raises F-score by 0.0123 [0.0062, 0.0187] and lowers Chamfer distance by 0.0157\,m [0.0076, 0.0236]. With matched cohorts (the three participants present in all ten scenes, and six participants matched within each of three acquisition blocks, with equal block weights as planned), it raises F-score by 0.0115 [0.0057, 0.0179] and lowers Chamfer distance by 0.0153\,m [0.0077, 0.0226]. Neither control measures data efficiency, because the cohorts change the evaluated recordings, not the training budgets. \Cref{app:posthoc} analyzes single recordings with the corrected generator (post hoc). Across scenes, the number of recordings (\cref{tab:app_rooms}) is not clearly related to the feature-replacement effect on \ours{} (Spearman correlation 0.14 for layout $r$ and $-0.31$ for the F-score at $a=2$, ten scenes, descriptive only).

\section{Per-scene results and atlas routes}
\label{app:corrected_scenes}

\Cref{tab:corrected_scenes} compares the generator and \ours{} in each held-out scene. Its means weight scenes equally, so they differ slightly from the route-pooled values of \cref{tab:main,tab:surface}. \Cref{tab:surface} gives every method at $a=1$ and $a=2$.

\begingroup
\makeatletter
\setlength{\@fptop}{0pt}
\setlength{\@fpbot}{0pt plus 1fil}
\makeatother

\begin{table}[t]
 \centering\scriptsize
 \caption{\textbf{\ours{} raises layout $r$ in every held-out scene, and its surface comparison with the generator depends on $a$.} At $a=2$ it lowers pixel MAE in every scene and Chamfer distance in eight, but its F-score is higher in only four. Scene means (equal weights, 324 routes) of the generator (G) and \ours{} (D). Pixel MAE and surfaces at $a=2$. D$-$G: mean difference, 95\% paired scene-bootstrap interval and scenes with D better. Bold: better mean. At $a=1$ the F-score and Chamfer distance of \ours{} are worse than those of the generator in every scene (\cref{tab:main_uncal}, \cref{app:extended}).}
 \label{tab:corrected_scenes}
 \setlength{\tabcolsep}{2.6pt}
 \begin{tabular*}{\linewidth}{@{\extracolsep{\fill}}ll rrrrrrrrrr r rcc@{}}
 \toprule
 & & \multicolumn{10}{c}{Held-out scene} & & & & D \\
 \cmidrule(lr){3-12}
 & & 1 & 2 & 3 & 4 & 5 & 6 & 7 & 8 & 9 & 10 & Mean & D$-$G & 95\% interval & better \\
 \midrule
Layout $r$\,\up & G & 0.307 & 0.336 & 0.314 & 0.283 & 0.343 & 0.277 & 0.285 & 0.275 & 0.262 & 0.324 & 0.301 & & & \\
 & D & 0.551 & 0.513 & 0.578 & 0.532 & 0.492 & 0.359 & 0.479 & 0.391 & 0.408 & 0.427 & \textbf{0.473} & $+$0.172 & [$+$0.135, $+$0.210] & 10/10 \\
\addlinespace[2pt]
Pixel MAE (m)\,\down & G & 1.419 & 1.493 & 1.577 & 1.434 & 1.529 & 1.559 & 1.410 & 1.512 & 1.668 & 1.572 & 1.517 & & & \\
 & D & 1.225 & 1.131 & 1.049 & 1.024 & 1.242 & 1.153 & 1.120 & 1.172 & 1.289 & 1.211 & \textbf{1.162} & $-$0.356 & [$-$0.408, $-$0.304] & 10/10 \\
\addlinespace[2pt]
F@25cm\,\up & G & 0.455 & 0.478 & 0.465 & 0.480 & 0.534 & 0.506 & 0.472 & 0.462 & 0.466 & 0.464 & 0.478 & & & \\
 & D & 0.447 & 0.507 & 0.530 & 0.534 & 0.496 & 0.504 & 0.481 & 0.456 & 0.441 & 0.443 & \textbf{0.484} & $+$0.006 & [$-$0.013, $+$0.026] & 4/10 \\
\addlinespace[2pt]
Chamfer (m)\,\down & G & 0.428 & 0.457 & 0.451 & 0.425 & 0.400 & 0.478 & 0.458 & 0.502 & 0.452 & 0.437 & 0.449 & & & \\
 & D & 0.446 & 0.378 & 0.339 & 0.349 & 0.412 & 0.415 & 0.406 & 0.437 & 0.435 & 0.433 & \textbf{0.405} & $-$0.044 & [$-$0.069, $-$0.018] & 8/10 \\
 \bottomrule
 \end{tabular*}
\end{table}

\paragraph{Route selection.}
The atlas has one route per scene: the lower-median route by the uncalibrated layout $r$ of \ours{}, with ties broken by route order. We froze this rule before rendering, so it does not depend on surface quality. \Cref{fig:surface_composite,fig:oblique_routes} instead show display routes: among the 144 where \ours{} and Brain-Diffuser + \ours{} layout both beat unmodified Brain-Diffuser in F-score, the highest by the F-score of \ours{} (\cref{fig:surface_composite}: clearest of the top five). In \cref{fig:surface_composite}, 39\% of the ground-truth points lie above 2.4\,m or outside the plotted area. As in the surface atlas (\cref{app:extended}), such points are hidden but scored (at most 3,907 predicted and 227 ground-truth points per panel). Unmodified Brain-Diffuser has slightly better mean surface scores at $a=1$ than \ours{} at $a=2$ (\cref{tab:surface}). \Cref{fig:depth_atlas_1} shows all eight depth views of the atlas routes of scenes 1 to 4 for \ours{}, unmodified Brain-Diffuser and Brain-Diffuser + \ours{} layout, with route layout $r$.

\begin{figure}[t]
 \centering
 \includegraphics[width=\linewidth]{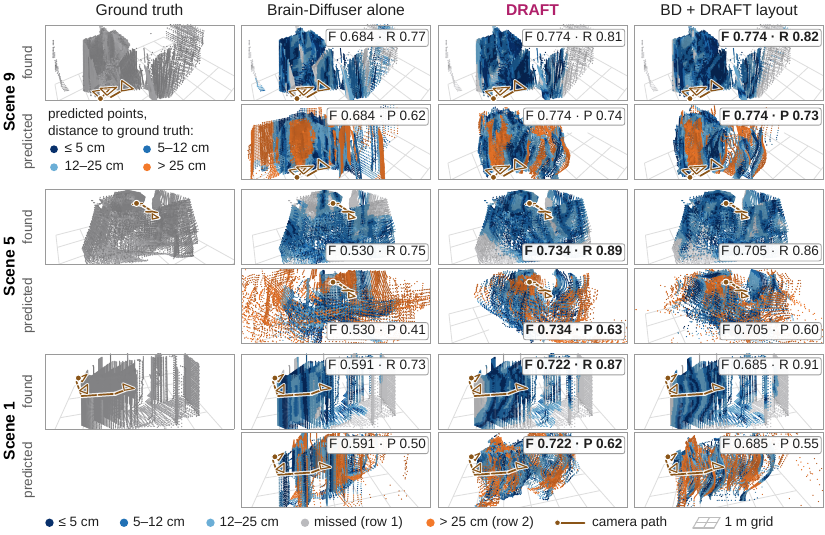}
 \caption{\textbf{Three routes chosen for display, not representative: predictions recover the room's extent and main wall positions, not clear walls or furniture.} Rule: among routes where \ours{} and Brain-Diffuser + \ours{} layout beat unmodified Brain-Diffuser (Brain-Diffuser alone) in F-score, highest \ours{} F-score first, one per scene, without the route of \cref{fig:surface_composite} (rule-chosen routes: \cref{tab:atlas_routes}). Brain-Diffuser alone: \cref{tab:baselines}. Row 1 (found): true points by distance to the nearest prediction (R: recall). Row 2 (predicted): predicted points by distance to the ground truth (P: precision). Points above 2.4\,m or outside the plotted area, and in row 2 stray points (fewer than 4 neighbors or clusters under 40 points, radii scaled with range), are hidden for display, but all points are scored. F: route F-score (bold: best of three).}
 \label{fig:oblique_routes}
\end{figure}

\paragraph{Failures.}
In the route of scene 5, fusion raises layout $r$ from 0.419 to 0.461, yet the F-score falls from 0.659 to 0.328. In scene 4 it instead rises from 0.536 to 0.686. One route can mislead either way, so we report route views (\cref{fig:depth_atlas_1}) and aggregate scores.

\begin{table}[t]
\centering\footnotesize
\caption{\textbf{On the ten atlas routes \ours{} improves the layout, but its F-score is the lowest of the three methods in seven routes.} Scores of the ten atlas routes, whose surface atlas is listed in \cref{app:extended}, all at $a=2$. G: generator, D: \ours{}, BD\,+\,L: Brain-Diffuser + \ours{} layout (surfaces only). Bold per row: best of three for surfaces. Mean: over the ten routes. Bottom: paired differences with 95\% route-bootstrap intervals and routes where D is better.}
\label{tab:atlas_routes}
\setlength{\tabcolsep}{4pt}
\begin{tabular}{@{}l rr@{\hspace{9pt}}rr@{\hspace{9pt}}rrr@{\hspace{9pt}}rrr@{}}
\toprule
 & \multicolumn{4}{c}{Depth map} & \multicolumn{6}{c}{Merged eight-view surface}\\
\cmidrule(lr){2-5}\cmidrule(lr){6-11}
 & \multicolumn{2}{c}{Layout $r$\,\up} & \multicolumn{2}{c}{\begin{tabular}[b]{@{}c@{}}Pixel MAE\\(m)\,\down\end{tabular}} & \multicolumn{3}{c}{F@25cm\,\up} & \multicolumn{3}{c}{Chamfer (m)\,\down}\\
\cmidrule(lr){2-3}\cmidrule(lr){4-5}\cmidrule(lr){6-8}\cmidrule(lr){9-11}
Scene & G & D & G & D & G & D & BD\,+\,L & G & D & BD\,+\,L\\
\midrule
1 & 0.536 & 0.563 & 0.988 & 0.856 & \textbf{0.574} & 0.541 & 0.488 & \textbf{0.277} & 0.279 & 0.304\\
2 & 0.356 & 0.500 & 1.258 & 0.846 & 0.560 & 0.550 & \textbf{0.606} & 0.426 & 0.299 & \textbf{0.252}\\
3 & 0.010 & 0.566 & 2.277 & 1.756 & \textbf{0.435} & 0.431 & 0.434 & 0.484 & 0.440 & \textbf{0.403}\\
4 & 0.435 & 0.583 & 1.221 & 0.705 & 0.536 & \textbf{0.686} & 0.635 & 0.399 & \textbf{0.238} & 0.259\\
5 & 0.419 & 0.461 & 1.288 & 1.349 & \textbf{0.659} & 0.328 & 0.415 & \textbf{0.291} & 0.448 & 0.371\\
6 & 0.065 & 0.400 & 1.623 & 1.188 & 0.442 & 0.468 & \textbf{0.480} & 0.591 & 0.466 & \textbf{0.460}\\
7 & 0.401 & 0.544 & 0.761 & 0.647 & \textbf{0.617} & 0.532 & 0.611 & 0.311 & 0.298 & \textbf{0.269}\\
8 & 0.359 & 0.400 & 1.362 & 1.109 & \textbf{0.621} & 0.556 & 0.583 & 0.369 & 0.392 & \textbf{0.321}\\
9 & 0.531 & 0.440 & 1.934 & 1.161 & 0.524 & 0.479 & \textbf{0.581} & 0.582 & 0.329 & \textbf{0.272}\\
10 & 0.077 & 0.478 & 1.897 & 1.436 & \textbf{0.485} & 0.397 & 0.432 & 0.466 & 0.509 & \textbf{0.418}\\
\midrule
Mean & 0.319 & 0.493 & 1.461 & 1.105 & \textbf{0.545} & 0.497 & 0.527 & 0.419 & 0.370 & \textbf{0.333}\\
 \midrule
D$-$G & \multicolumn{2}{c}{$+$0.175} & \multicolumn{2}{c}{$-$0.356} & \multicolumn{3}{c}{$-$0.048} & \multicolumn{3}{c}{$-$0.050}\\
95\% interval & \multicolumn{2}{c}{\scriptsize [$+$0.065, $+$0.297]} & \multicolumn{2}{c}{\scriptsize [$-$0.497, $-$0.212]} & \multicolumn{3}{c}{\scriptsize [$-$0.125, $+$0.016]} & \multicolumn{3}{c}{\scriptsize [$-$0.119, $+$0.019]}\\
D better & \multicolumn{2}{c}{9/10} & \multicolumn{2}{c}{9/10} & \multicolumn{3}{c}{2/10} & \multicolumn{3}{c}{6/10}\\[2pt]
D$-$(BD\,+\,L) & & & & & \multicolumn{3}{c}{$-$0.030} & \multicolumn{3}{c}{$+$0.037}\\
95\% interval & & & & & \multicolumn{3}{c}{\scriptsize [$-$0.061, $+$0.003]} & \multicolumn{3}{c}{\scriptsize [$+$0.012, $+$0.060]}\\
D better & & & & & \multicolumn{3}{c}{2/10} & \multicolumn{3}{c}{2/10}\\
\bottomrule
\end{tabular}
\end{table}

\begin{figure}[p]
 \centering
 \includegraphics[width=\linewidth]{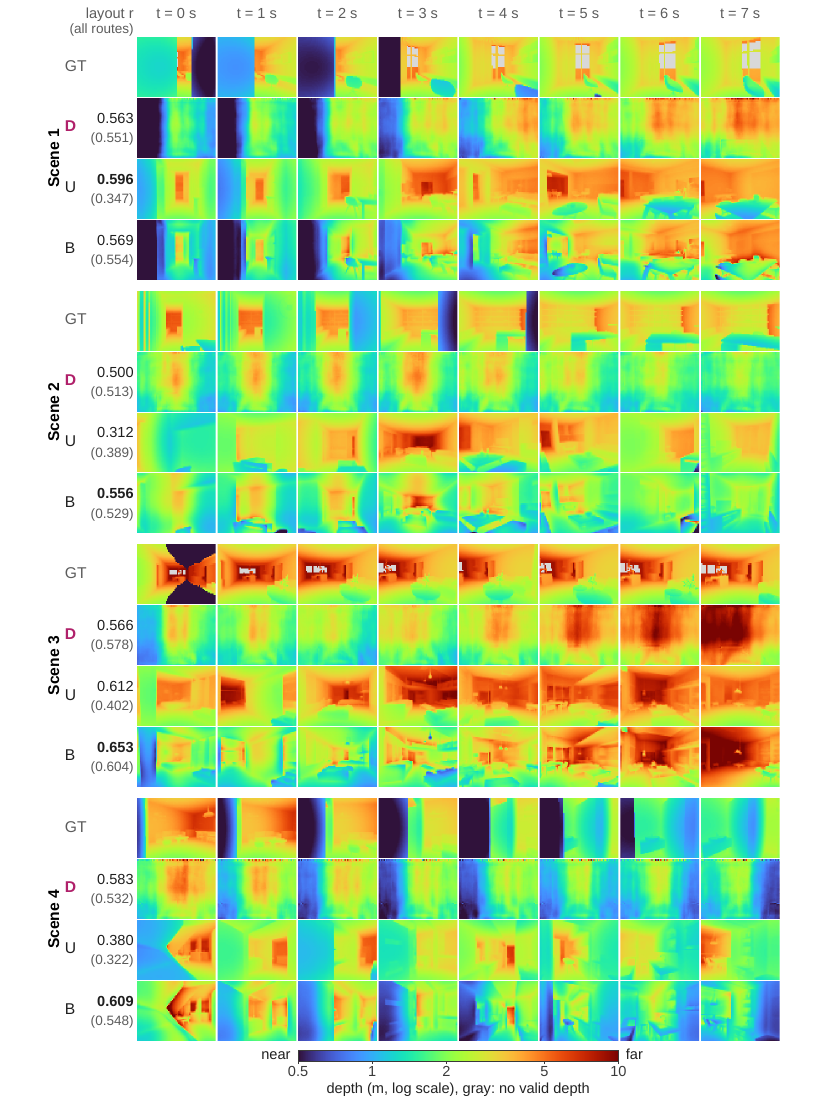}
 \caption{\textbf{Depth atlas, scenes 1 to 4, failures included: on the ten rule-chosen routes, Brain-Diffuser + \ours{} layout has the highest route layout $r$ in six scenes, \ours{} in three and unmodified Brain-Diffuser in one.} All eight views, 1\,s apart, of the rule-chosen routes of \cref{tab:atlas_routes} (\cref{fig:teaser} is hand-picked). Rows: furnished ground truth (GT, gray: invalid), \ours{} (D, $a=2$), unmodified Brain-Diffuser (U, calibrated DA V2 depth, $a=1$) and Brain-Diffuser + \ours{} layout (B, $a=2$). Numbers: route layout $r$ (bold: best of three) and the scene's value over all routes (in parentheses). Means over these ten and all 324 routes: 0.493 and 0.480 (D), 0.236 and 0.324 (U), 0.509 and 0.493 (B). Scenes 5 to 10: \cref{app:extended}.}
 \label{fig:depth_atlas_1}
\end{figure}
\endgroup

\section{Surface analyses}
\label{app:surfaces}
This section tests three explanations of the surface scores: depth range (\cref{app:aligned}), agreement across views (\cref{app:horizon}) and pose drift (\cref{app:registration_drift}).

\subsection{Depth scale}
\label{app:aligned}
To test whether the surface gap between methods is only a matter of depth range, we align each predicted view to its ground truth in log depth before lifting. Median alignment matches the medians~\citep{zhou2017unsupervised}. The log-affine alignment also fits a slope, so it undoes range calibration as well~\citep[compare][]{ranftl2022towards}. A least-squares slope shrinks noisy predictions toward the mean. This fit therefore partly measures per-view correlation again, and even the fixed template reaches F-score 0.575 under it. Both alignments use test depth and are post hoc. We also report three unaligned pixel metrics: the scale-invariant log error (SI-log, $\times100$)~\citep{eigen2014depth}, the mean absolute relative error (AbsRel), and the share of pixels within a factor of 1.25 of the true depth ($\delta{<}1.25$).

Under median alignment, the \ours{} layout still helps (\cref{tab:aligned}, intervals in panel b). \ours{} is above Brain-Diffuser in F-score by 0.020 and above the trajectory head by 0.008, each in nine scenes, and above the generator and MindEye in all ten. Brain-Diffuser + \ours{} layout is better still. It is above \ours{} in nine scenes and has a lower Chamfer distance in ten. Applied to the depth of Brain-Diffuser, the \ours{} layout beats the trajectory layout in F-score, Chamfer distance and SI-log. Under the log-affine fit the two layouts tie.

Under the log-affine fit, \ours{} has the highest F-score but ties the trajectory head and Brain-Diffuser + \ours{} layout, and the latter has the lower log-affine Chamfer distance. It is above Brain-Diffuser by 0.024 and above MindEye by 0.031, in all ten scenes. Its margins over the generator, the template and linear depth are 0.034, 0.050 and 0.019 (from unrounded means).

Without alignment, \ours{} has a lower SI-log than the trajectory head, generator, Brain-Diffuser and MindEye in all ten scenes ($p=0.002$ each). Brain-Diffuser + \ours{} layout has a higher SI-log than \ours{} in all ten scenes ($p=0.002$), by only 0.5. Linear depth ties \ours{} (lower in five scenes, $p=0.29$). Image decoders have a higher $\delta{<}1.25$ (nine scenes each).

\begin{table}[ht]
  \centering
  \caption{\textbf{Once per-view depth scale is removed by alignment, the methods with a \ours{} or trajectory layout score best.} (a) Means over ten held-out scenes at $a=1$. Alignment uses test depth (post hoc). F: F-score at 25\,cm. SI-log: $\times100$. Bold, underline: best and second best per column, ties share a mark. (b) Paired differences, 95\% paired scene-bootstrap intervals, scenes favoring the first method and sign-flip $p$ (\cref{app:signflip}). $^\dagger$First minus second. BD: Brain-Diffuser.}
  \label{tab:aligned}
  \footnotesize
  \renewcommand{\arraystretch}{0.88}
  \setlength{\tabcolsep}{3.0pt}
  \begin{tabular}{@{}lrr@{\hspace{8pt}}rr@{\hspace{8pt}}rrr@{}}
    \toprule
    \textit{(a)} & \multicolumn{2}{c}{Median-aligned} & \multicolumn{2}{c}{Log-affine fit} & \multicolumn{3}{c}{Unaligned pixels} \\
    \cmidrule(r){2-3}\cmidrule(lr){4-5}\cmidrule(l){6-8}
    Method & F\,\up & Chamfer (m)\,\down & F\,\up & Chamfer (m)\,\down & SI-log\,\down & AbsRel\,\down & $\delta{<}1.25$\,\up \\
    \midrule
    Fixed template & 0.569 & 0.331 & 0.575 & 0.319 & 52.8 & 1.150 & 0.281 \\
    Generator & 0.548 & 0.340 & 0.592 & 0.316 & 54.9 & 0.851 & 0.265 \\
    Trajectory head & 0.582 & 0.324 & 0.622 & 0.280 & 50.6 & 0.857 & 0.223 \\
    Linear depth & 0.581 & 0.322 & 0.607 & 0.290 & \second{50.1} & 0.825 & 0.234 \\
    Brain-Diffuser & 0.571 & 0.328 & 0.601 & 0.291 & 57.3 & 1.064 & \best{0.291} \\
    MindEye & 0.558 & 0.345 & 0.594 & 0.297 & 59.5 & 1.030 & \second{0.283} \\
    \midrule
    BD + trajectory layout & \second{0.598} & \second{0.298} & 0.622 & \second{0.271} & 51.1 & 0.865 & 0.219 \\
    BD + \ours{} layout & \best{0.608} & \best{0.291} & \second{0.623} & \best{0.268} & \second{50.1} & \second{0.821} & 0.244 \\
    \midrule
    \ours{} & 0.591 & 0.319 & \best{0.625} & 0.277 & \best{49.6} & \best{0.813} & 0.248 \\
    \midrule
    \ours{} $-$ strongest other & $-$0.017 & $+$0.028 & $+$0.002 & $+$0.008 & $-$0.5 & $-$0.008 & $-$0.043 \\
    \quad 95\% interval, lower & $-$0.022 & $+$0.021 & $-$0.002 & $+$0.004 & $-$1.3 & $-$0.010 & $-$0.074 \\
    \quad 95\% interval, upper & $-$0.011 & $+$0.034 & $+$0.006 & $+$0.013 & $+$0.3 & $-$0.007 & $-$0.013 \\
    \bottomrule
  \end{tabular}

  \vspace{2.5pt}
  \setlength{\tabcolsep}{3.5pt}
  \begin{tabular*}{\linewidth}{@{\extracolsep{\fill}}lllcc@{}}
    \multicolumn{5}{@{}l}{\textit{(b) Paired contrasts (Chamfer and SI-log: reduction, positive favors the first, blank: not reported)}} \\
    \toprule
    Contrast & Metric & Difference [95\% CI] & Scenes & $p$ \\
    \midrule
    \multicolumn{5}{@{}l}{\emph{Median-aligned}} \\
    \ours{} vs Brain-Diffuser & F & $+0.020$ [$+0.009$, $+0.030$] & 9 & 0.006 \\
    \ours{} vs trajectory head & F & $+0.008$ [$+0.004$, $+0.012$] & 9 & \\
    BD + \ours{} layout vs \ours{} & F & $+0.017$ [$+0.011$, $+0.022$] & 9 & \\
    BD + \ours{} vs BD + trajectory layout & F & $+0.010$ [$+0.006$, $+0.014$] & 10 & 0.002 \\
     & Chamfer (m) & $+0.007$ [$+0.003$, $+0.010$] & 8 & \\
    \multicolumn{5}{@{}l}{\emph{Log-affine fit}} \\
    BD + \ours{} vs BD + trajectory layout & F & $+0.001$ [$-0.003$, $+0.005$] & & \\
     & Chamfer (m)$^\dagger$ & $-0.003$ [$-0.006$, $+0.001$] & & \\
    \ours{} vs Brain-Diffuser & F & $+0.024$ [$+0.017$, $+0.032$] & 10 & 0.002 \\
    \ours{} vs MindEye & F & $+0.031$ [$+0.024$, $+0.037$] & 10 & \\
    \ours{} vs trajectory head & F & $+0.003$ [$-0.002$, $+0.008$] & 6 & 0.24 \\
    \ours{} vs BD + \ours{} layout & F & $+0.002$ [$-0.002$, $+0.006$] & 7 & 0.43 \\
    BD + \ours{} layout vs \ours{} & Chamfer (m) & $+0.008$ [$+0.004$, $+0.013$] & 8 & \\
    \multicolumn{5}{@{}l}{\emph{Unaligned}} \\
    BD + \ours{} vs BD + trajectory layout & SI-log & $+0.98$ [$+0.62$, $+1.41$] & 10 & 0.002 \\
    \ours{} vs Brain-Diffuser & SI-log & $+7.68$ [$+7.09$, $+8.32$] & 10 & 0.002 \\
    \ours{} vs MindEye & SI-log & $+9.87$ [$+9.05$, $+10.62$] & 10 & 0.002 \\
    \bottomrule
  \end{tabular*}

\end{table}

\subsection{Agreement across views}
\label{app:horizon}

We test whether a route's views agree by comparing their errors where two views see one point.

\paragraph{Change in 3D error between views.}
The pixel pairs come from ground truth alone, so every method is scored on the same pairs. The nine eight-view routes of a video give 72 views. For views $t$ and $t+\ell$, we project each valid ground-truth point into the second camera. We keep the pair only if its projected depth is positive and the target pixel is valid. The two radial depths must also agree within $\max(0.10\,\mathrm m,0.02r)$ for ground-truth radial depth $r$, and the two ground-truth points must lie within 0.10\,m. With predicted points $\hat p$ and true points $p$ of a matched pair, the change in 3D error is
\begin{equation}
 E_\ell=\operatorname{mean}\left\| (\hat p_t-p_t)-(\hat p_{t+\ell}-p_{t+\ell})\right\|_2.
 \label{eq:error_disagreement}
\end{equation}
At a lag of one second, matched ground-truth points lie 0.038\,m apart on average. We average pairs within a view pair, then view pairs within a video, videos within a scene and scenes equally.

At $a=1$, \ours{} has a smaller change than the generator and Brain-Diffuser in all ten scenes, by 0.213\,m [0.190, 0.236] and 0.543\,m [0.525, 0.562]. Linear depth with the \ours{} layout has the smallest change, and MindEye was not scored. The gain over Brain-Diffuser shrinks as the time between views grows (lag analysis in \cref{app:extended}). But a small change in 3D error is not a small error. An error shared by both views cancels, so a surface misplaced the same way in every view, such as a constant template (not scored), also has a small change. The change must therefore be read with the surface scores. In the routes drawn view by view (\cref{app:extended}), every method misplaces the walls, and no method puts a wall seen by consecutive views at one place.

Still, \ours{} has a smaller change than Brain-Diffuser with the \ours{} layout in all ten scenes at both factors, by 0.186\,m [0.168, 0.205] at $a=2$. Linear depth with the same layout has a similar or smaller change than \ours{} (0.581 against 0.627\,m at $a=2$, with overlapping intervals), so a small change is not specific to our generator. \ours{} itself is second best in three of the four columns of \cref{tab:horizon_ci} and never best. For \ours{} at $a=2$, the one-second change is 0.623\,m within a route and 0.653\,m across route boundaries. These means come from different view pairs, so they do not measure a boundary effect.

\paragraph{Fixed temporal smoothing.}
\label{app:temporal_control}
Plain smoothing over time does not help linear depth, a view-by-view decoder, in the same way. A fixed three-view filter (weights 0.25, 0.5 and 0.25 in log depth, within each route) changes the F-score by $-0.0024$ and raises its Chamfer distance by 0.0034\,m [0.0027, 0.0043]. With the trajectory layout the changes are $-0.0085$ and 0.0159\,m [0.0138, 0.0179]. Chamfer distance is worse in all ten scenes in both cases.

\begin{table}[ht]
 \centering\footnotesize
 \caption{\textbf{The \ours{} layout lowers the change in 3D error between views for every source at both factors, but at $a=1$ it lowers the F-score of \ours{} and Brain-Diffuser.} Eight-view F-score and one-second change in 3D error $E_1$ of \cref{eq:error_disagreement} over the 72 views of each video. Ten-scene means with 95\% scene-bootstrap intervals (paired for differences). Bold, underline: best and second best. Bottom: with minus without the \ours{} layout. BD: Brain-Diffuser.}
 \label{tab:horizon_ci}
 \setlength{\tabcolsep}{3.5pt}
 \begin{tabular}{@{}lcccc@{}}
 \toprule
 & \multicolumn{2}{c}{F@25cm, eight views\,\up} & \multicolumn{2}{c}{Change in 3D error (m), 1\,s lag\,\down} \\
 \cmidrule(lr){2-3}\cmidrule(lr){4-5}
 Prediction source & $a=1$ & $a=2$ & $a=1$ & $a=2$ \\
 \midrule
 \multicolumn{5}{@{}l}{\emph{Current models}}\\
 Generator & \underline{0.469} \ci{0.442}{0.493} & 0.456 \ci{0.435}{0.477} & 0.585 \ci{0.561}{0.613} & 1.227 \ci{1.170}{1.292} \\
 \ours{} & 0.418 \ci{0.391}{0.445} & \underline{0.467} \ci{0.435}{0.494} & \underline{0.372} \ci{0.347}{0.398} & \underline{0.627} \ci{0.577}{0.681} \\
 \addlinespace[2pt]
 \multicolumn{5}{@{}l}{\emph{Decoders}}\\
 Brain-Diffuser & \textbf{0.492} \ci{0.467}{0.516} & 0.411 \ci{0.398}{0.427} & 0.915 \ci{0.887}{0.944} & 2.404 \ci{2.314}{2.498} \\
 BD + \ours{} layout & 0.436 \ci{0.408}{0.463} & \textbf{0.481} \ci{0.450}{0.509} & 0.424 \ci{0.397}{0.453} & 0.813 \ci{0.751}{0.880} \\
 Linear depth & 0.386 \ci{0.351}{0.419} & 0.458 \ci{0.440}{0.477} & 0.378 \ci{0.355}{0.402} & 0.660 \ci{0.604}{0.721} \\
 Linear depth + \ours{} layout & 0.393 \ci{0.364}{0.422} & 0.446 \ci{0.414}{0.473} & \textbf{0.353} \ci{0.328}{0.380} & \textbf{0.581} \ci{0.531}{0.637} \\
 \midrule
 \multicolumn{5}{@{}l}{\emph{Paired differences, with minus without the \ours{} layout}}\\
 \ours{} & $-$0.050 & $+$0.011 & $-$0.213 & $-$0.600 \\
 \quad minus generator & \ci{$-$0.078}{$-$0.023} & \ci{$-$0.024}{$+$0.043} & \ci{$-$0.236}{$-$0.190} & \ci{$-$0.671}{$-$0.524} \\
 BD + \ours{} layout & $-$0.056 & $+$0.070 & $-$0.490 & $-$1.591 \\
 \quad minus BD & \ci{$-$0.095}{$-$0.020} & \ci{$+$0.041}{$+$0.097} & \ci{$-$0.509}{$-$0.471} & \ci{$-$1.672}{$-$1.515} \\
 Linear depth + \ours{} layout & $+$0.008 & $-$0.012 & $-$0.025 & $-$0.079 \\
 \quad minus linear depth & \ci{$-$0.013}{$+$0.029} & \ci{$-$0.039}{$+$0.014} & \ci{$-$0.039}{$-$0.011} & \ci{$-$0.122}{$-$0.043} \\
 \bottomrule
 \end{tabular}
\end{table}

\subsection{Pose drift during registration}
\label{app:registration_drift}
Camera poses of a new recording may be noisy, so we test how pose drift changes the surfaces, at the registration step only. The predicted depth is fixed and uncalibrated ($a=1$). Only the poses that lift predicted points and merge them on the 5\,cm voxel grid drift, and ground truth keeps the clean poses. The drift is a horizontal translation (5, 10 or 20\,cm), a yaw (1, 2 or 5 degrees) or both (10\,cm with 2 degrees). It grows linearly from zero at the first view to its final value at the eighth, with five fixed seeds per route shared across levels and methods. Each method is compared with its own surface under clean poses.

The primary measure, fixed in advance, is the change in the F-score of \ours{} under joint drift. It is $-0.026$ percentage points [$-0.100$, $+0.039$], an interval that includes zero but does not establish equivalence. The Chamfer distance of \ours{} rises by 0.35\,mm [0.02, 0.67]. The largest effect, 20\,cm of translation for the generator, lowers the F-score by 0.165 percentage points and raises the Chamfer distance by 1.57\,mm. Every mean change in every condition is below 0.2 percentage points in F-score, precision and recall and below 2\,mm in Chamfer distance (all conditions omitted; \cref{app:extended}). These results do not show that upstream pose estimation or the full range-calibrated pipeline ($a=2$) is robust, and they do not rank the two methods. \Cref{app:posthoc} perturbs the camera-path inputs instead (post hoc).

\section{fMRI-free references, generator detail and range calibration}
\label{app:floors}
The range-calibration factor $a$ is that of \cref{sec:fusion}, and $a=1$ means no calibration.

\paragraph{fMRI-free references.}
The \textbf{fixed template} is the per-pixel median ground-truth log depth of the nine training scenes of each fold. It is the same map for every held-out view, and we lift and score it like any prediction. Route-pooled, it reaches layout $r$ 0.324, row order 0.581, left/right 0.509 and pixel MAE 1.188\,m. The \textbf{per-second median} instead takes, for each view, the median training-scene log depth at the same video second. It reaches layout $r$ 0.241 and pixel MAE 1.254\,m. Its layout $r$ is below that of the fixed template by 0.081 [0.045, 0.114] (nine scenes). So knowing the video second does not help. \ours{} is above the per-second median in layout $r$ in all ten scenes (\cref{tab:contrasts}). Its pixel MAE is lower by 0.128\,m [0.028, 0.225] (eight scenes). We did not compute a readout from the constant and linear-trend regressors alone. The ten-scene mean layout $r$ gain over the fixed template is $+0.144$ for \ours{} and $+0.157$ for Brain-Diffuser + \ours{} layout. It is $+0.109$ for the trajectory head and $+0.099$ for linear depth and the DREAM depth branch. Brain-Diffuser, MindEye and the generator stay at or below the template (\cref{sec:main}).

The \textbf{wrong-scene control} asks how much of a surface score is specific to the scene. We lift the predictions made for the next held-out scene at the poses of the evaluated routes and score them against the evaluated scene. The correct-scene score minus this score is the scene-specific gain (\cref{tab:floors}). Both references are post hoc.

\begin{table}[ht]
  \centering
  \caption{\textbf{Wrong-scene predictions keep most of the F-score.} Upper rows: the fixed template scored as a prediction. Lower rows: predictions for the next held-out scene lifted at the evaluated poses (post hoc), and the gain of the correct scene, with its 95\% paired scene-bootstrap interval and the number of scenes (of ten) with a positive gain. Ten-scene means. --: not applicable, since the template is the same for every scene. Brain-Diffuser and MindEye at their best factor, $a=1$.}
  \label{tab:floors}
  \footnotesize
  \def\flhd#1{\begin{tabular}[b]{@{}c@{}}#1\end{tabular}}
  \setlength{\tabcolsep}{2.6pt}
  \begin{tabular}{@{}lcrrrrrlr@{}}
    \toprule
    & & \multicolumn{4}{c}{Scored against the evaluated scene} & \multicolumn{1}{c}{Correct} & \multicolumn{2}{c}{Correct minus wrong scene} \\
    \cmidrule(lr){3-6}\cmidrule(lr){7-7}\cmidrule(l){8-9}
    Prediction & $a$ & F@25cm\,\up & Precision\,\up & Recall\,\up & \flhd{Chamfer\\(m)\,\down} & F@25cm\,\up & \multicolumn{1}{c}{\flhd{F gain\,\up\\{}[95\% CI]}} & \flhd{Scenes\\(of 10)} \\
    \midrule
    \multicolumn{9}{@{}l}{\emph{Fixed template}} \\
    Fixed template & 1 & 0.391 & 0.399 & 0.405 & 0.524 & \multicolumn{1}{c}{--} & \multicolumn{1}{c}{--} & \multicolumn{1}{c}{--} \\
     & 2 & 0.437 & 0.390 & 0.519 & 0.437 & \multicolumn{1}{c}{--} & \multicolumn{1}{c}{--} & \multicolumn{1}{c}{--} \\
    \midrule
    \multicolumn{9}{@{}l}{\emph{Wrong-scene control: predictions for the next held-out scene}} \\
    \ours{} & 1 & 0.379 & 0.379 & 0.403 & 0.546 & 0.416 & $+0.037$ [0.011, 0.063] & 6 \\
     & 2 & 0.415 & 0.356 & 0.527 & 0.492 & 0.484 & $+0.069$ [0.042, 0.098] & 10 \\
    Generator alone & 1 & 0.444 & 0.402 & 0.518 & 0.467 & 0.477 & $+0.033$ [0.017, 0.049] & 9 \\
     & 2 & 0.443 & 0.336 & 0.683 & 0.502 & 0.478 & $+0.035$ [0.019, 0.052] & 10 \\
    Brain-Diffuser & 1 & 0.433 & 0.363 & 0.557 & 0.464 & 0.492 & $+0.059$ [0.052, 0.065] & 10 \\
    MindEye & 1 & 0.439 & 0.375 & 0.553 & 0.451 & 0.494 & $+0.055$ [0.040, 0.069] & 10 \\
    \bottomrule
  \end{tabular}
\end{table}

\paragraph{What the generator adds.}
To test whether the generator adds more than the layout, we remove its detail (the high-frequency part) from the \ours{} depth, leaving only the layout, or replace it with the detail of the fixed template (all rows omitted; \cref{app:extended}). At $a=2$, removing the detail lowers the F-score in all ten scenes. \ours{} minus the layout without detail is $+0.042$ [0.036, 0.047] in F-score and $-0.049$\,m [$-0.057$, $-0.040$] in Chamfer distance. The template detail recovers part of this loss, but \ours{} stays ahead of it by $+0.025$ [0.020, 0.030] in F-score. Layout $r$ changes by at most 0.006. Most of the detail's energy is one map shared by all scenes and views (share of detail energy in that map 0.551 against 0.007 in the ground truth, a difference of $+0.544$ [0.479, 0.648], post hoc), and it barely tracks the true detail (correlation 0.026 [0.021, 0.031]). The generator therefore acts mostly as a learned prior of indoor depth detail. The vertical stripes of $\mD$ in \cref{fig:pipeline} belong to this shared map.

\paragraph{Guiding other decoders with a layout.}\label{app:layout_guidance}
The ``+ path'' and ``+ \ours{}'' columns of \cref{tab:baselines} combine a layout with the depth of another decoder. They apply the low-frequency replacement of \cref{eq:lowfreq} to that depth instead of the generator's, using the trajectory layout or the \ours{} layout. The replacement keeps the fine detail of the decoder's depth and takes its broad structure from the layout. For the two planned decoders, the depth is the calibrated DA V2 depth of each Brain-Diffuser image or the linear depth prediction. It reuses the implementation, $\sigma=4$ pixels, layouts and range calibration of \ours{} and has no fitted parameters. For these two decoders, all combined predictions were built and frozen under a written plan before any of them was scored. The other five decoders were added post hoc (\cref{app:transfer}). Under the \ours{} layout at $a=2$, the detail of Brain-Diffuser raises recall over that of our generator by 0.071. It lowers precision by 0.011. Both differences hold in all ten scenes (\cref{tab:prec_recall}b).
Without the DA V2 scale calibration, the F-scores of Brain-Diffuser and MindEye fall to 0.242 and 0.258. Their Chamfer distances rise to 0.973 and 0.847\,m. Their surface scores therefore depend on the depth scale learned from the training scenes.

\paragraph{How $a=2$ was chosen.}\label{app:range_sensitivity}
We fixed $a=2$ after an earlier sweep over the historical models (before the loss correction) on these scenes. That sweep reused predictions of other outer folds, so the choice was descriptive and not nested. The current sweep and the nested selection below test it.

\paragraph{Range calibration ablation.}\label{app:range_ablation}
\Cref{tab:main_uncal} scores the layouts of \cref{tab:main} and the main methods of \cref{tab:baselines} with and without range calibration (the uncalibrated planned combinations and contrasts are summarized in \cref{app:extended}). Without calibration, the generator alone has the best F-score and Chamfer distance of the five layouts. Calibration raises the F-score of every layout except the generator, whose F-score barely changes, and it raises every pixel MAE. \ours{} lowers pixel MAE relative to the trajectory head in all ten scenes without calibration (by 0.077\,m [0.046, 0.108]) and in nine with it (\cref{tab:contrasts}). Its wrong-scene gain in \cref{fig:surface_composite}f is 0.069 with calibration and 0.037 without it (\cref{tab:floors}).

\begin{table}[ht]
  \centering
  \caption{\textbf{Without range calibration ($a=1$), fusion lowers the F-score and raises the Chamfer distance. Calibration raises the F-score of every layout except the generator's and lowers it for the image decoders.} C: EVC, M: camera motion, T: free-space features. Layout $r$ and pixel MAE: 324-route means, F-score and Chamfer distance: ten-scene means, as in \cref{tab:main}. Bold, underline: best and second best current method. Small lines: $a=2$ minus $a=1$, 95\% paired scene-bootstrap interval. \ours{} minus generator: ten-scene means of per-scene differences ($^\ast$: interval excludes zero). $^\dagger$Fixed after earlier results on these scenes. BD: Brain-Diffuser.}
  \label{tab:main_uncal}\label{tab:surface}
  \footnotesize
  \setlength{\tabcolsep}{3.2pt}
  \def\dg#1{{\scriptsize#1}}\def\sa{\rlap{\smash{$^{\textstyle\ast}$}}}
  \def\dl#1#2{\multicolumn{2}{c}{\scriptsize #1}&\multicolumn{2}{c}{\scriptsize #2}\\[3pt]}
  \def\sk{\addlinespace[-1.5pt]\multicolumn{4}{@{}l}{}&}
  \begin{tabular}{@{}lccccccc@{}}
    \toprule
    & & \multicolumn{2}{c}{Pixel MAE (m)\,\down} & \multicolumn{2}{c}{F@25cm\,\up} & \multicolumn{2}{c}{Chamfer (m)\,\down} \\
    \cmidrule(lr){3-4}\cmidrule(lr){5-6}\cmidrule(l){7-8}
    Method (inputs) & Layout $r$\,\up & $a{=}1$ & $a{=}2^\dagger$ & $a{=}1$ & $a{=}2^\dagger$ & $a{=}1$ & $a{=}2^\dagger$ \\
    \midrule
    Generator alone (none) & 0.303 & 1.168 & 1.507 & 0.477 & 0.478 & 0.426 & 0.449 \\
    \sk\dl{$+$0.002 [$-$0.012, $+$0.013]}{$+$0.023 [$-$0.002, $+$0.047]}
    Readout, \threegrid{} target (C, M) & 0.440 & \best{1.097} & \second{1.176} & 0.432 & \second{0.485} & 0.478 & 0.417 \\
    Trajectory head (M) & 0.390 & 1.217 & 1.264 & 0.383 & 0.447 & 0.535 & 0.443 \\
    Trajectory head (M, T) & 0.446 & 1.185 & 1.252 & 0.397 & 0.465 & 0.514 & 0.427 \\
    \sk\dl{$+$0.068 [$+$0.044, $+$0.088]}{$-$0.087 [$-$0.123, $-$0.050]}
    Fusion head, \ours{} (C, M, T) & \second{0.480} & \second{1.107} & \best{1.153} & 0.416 & 0.484 & 0.491 & \second{0.405} \\
    \sk\dl{$+$0.068 [$+$0.045, $+$0.090]}{$-$0.086 [$-$0.126, $-$0.044]}
    Brain-Diffuser & 0.324 & 1.212 & 2.326 & \second{0.492} & 0.424 & \second{0.389} & 0.865 \\
    \sk\dl{$-$0.068 [$-$0.082, $-$0.057]}{$+$0.477 [$+$0.420, $+$0.534]}
    MindEye & 0.270 & 1.209 & 2.119 & \best{0.494} & 0.416 & \best{0.381} & 0.744 \\
    \sk\dl{$-$0.078 [$-$0.098, $-$0.059]}{$+$0.362 [$+$0.315, $+$0.404]}
    BD + \ours{} layout & \best{0.493} & 1.112 & 1.197 & 0.434 & \best{0.499} & 0.444 & \best{0.377} \\
    \sk\dl{$+$0.064 [$+$0.041, $+$0.087]}{$-$0.067 [$-$0.110, $-$0.020]}
    Fixed template & 0.324 & 1.188 & 1.239 & 0.391 & 0.437 & 0.524 & 0.437 \\
    \sk\dl{$+$0.047 [$+$0.033, $+$0.060]}{$-$0.087 [$-$0.104, $-$0.071]}
    \midrule
    \ours{} minus generator & \dg{+0.172\sa} & \dg{\textminus0.055\sa} & \dg{\textminus0.356\sa} & \dg{\textminus0.061\sa} & \dg{+0.006} & \dg{+0.065\sa} & \dg{\textminus0.044\sa} \\
    Scenes with \ours{} better & \dg{10/10} & \dg{7/10} & \dg{10/10} & \dg{0/10} & \dg{4/10} & \dg{0/10} & \dg{8/10} \\
    \midrule
    \multicolumn{8}{@{}l}{\emph{Historical, before the correction (not ranked)}} \\
    Generator & 0.369 & 1.219 & 1.435 & 0.426 & 0.463 & 0.513 & 0.518 \\
    \sk\dl{$+$0.037 [$+$0.023, $+$0.051]}{$+$0.005 [$-$0.035, $+$0.046]}
    \ours{} & 0.486 & 1.118 & 1.178 & 0.409 & 0.462 & 0.505 & 0.433 \\
    \sk\dl{$+$0.052 [$+$0.032, $+$0.071]}{$-$0.072 [$-$0.110, $-$0.032]}
    \bottomrule
  \end{tabular}
\end{table}

\paragraph{Sweep and nested selection.}
\Cref{tab:a_sweep} scores each method at $a$ from 1 to 3. Chosen on the held-out scenes, the best factor is $a=2$ for \ours{}, 1.5 for the generator alone and 1 for Brain-Diffuser and MindEye. At these factors \ours{} is not better than the generator (\cref{tab:nested_a}). Its F-score is lower by 0.016 and its Chamfer distance higher by 0.014\,m, both within noise, and the generator has the higher recall in nine scenes. MindEye has a lower Chamfer distance than \ours{}, by 0.024\,m. Every fMRI method at its best factor still beats the template in F-score (\ours{} by 0.042, nine scenes). The generator loses precision as $a$ grows, so part of the higher precision of \ours{} at $a=2$ reflects the generator's over-calibration.

To avoid choosing $a$ on the scored scene, we pick $a$ for each held-out scene by the mean score of the other nine. By Chamfer distance, this picks the same factor as the sweep for every method except Brain-Diffuser + \ours{} layout. By F-score, \ours{} picks $a=2.5$ in five folds, because $a=2$ and 2.5 tie at 0.484. Its Chamfer distance then rises to 0.439\,m. Its F-score falls below those of the generator and MindEye, and its Chamfer distance is worse than every image decoder's (\cref{tab:nested_a}b).

\paragraph{Brain-Diffuser + \ours{} layout at its best factor.}
At its best factor, this combination is ahead of \ours{}. Its F-score is higher by 0.015, in nine scenes (\cref{tab:baselines,tab:baselines_ci}). Its Chamfer distance is lower by 0.028\,m [0.012, 0.042], in eight scenes. Against the unmodified decoders at $a=1$, it is within noise. Its F-score is higher than those of Brain-Diffuser and MindEye by 0.006 [$-0.009$, 0.022] and 0.004 [$-0.007$, 0.016]. Its Chamfer distance is lower than theirs by 0.012\,m [$-0.012$, 0.032] and 0.004\,m [$-0.020$, 0.023], respectively. The F-score favors it in five scenes against each decoder, and the Chamfer distance in seven.

\begin{table}[ht]
  \centering
  \caption{\textbf{At each method's best factor, \ours{} surfaces are no better than those of the generator or the image decoders.} (a) F-score and Chamfer distance for $a$ from 1 to 3 (bold: best $a$, chosen on the held-out scenes, paired differences in \cref{tab:nested_a}). (b) Precision and recall at $a=2$ and at the best $a$, and \ours{} minus the row (--: reference, bold: best at $a=2$, the $a=1$ block is summarized in \cref{app:extended}). Ten-scene means, 95\% paired scene-bootstrap intervals.}
  \label{tab:a_sweep}
  \label{tab:prec_recall}
  \footnotesize
  \def\flhd#1{\begin{tabular}[b]{@{}c@{}}#1\end{tabular}}
  \setlength{\tabcolsep}{9pt}
  \begin{tabular*}{\linewidth}{@{\extracolsep{\fill}}l*{5}{r}@{}}
    \multicolumn{6}{@{}l}{\textit{(a) Calibration-factor sweep}} \\
    \toprule
    & \multicolumn{5}{c}{Calibration factor $a$} \\
    \cmidrule(l){2-6}
    Method & 1 & 1.5 & 2 & 2.5 & 3 \\
    \midrule
    \multicolumn{6}{@{}l}{\emph{F@25cm\,\up}} \\
    \ours{} & 0.416 & 0.464 & \best{0.484} & \best{0.484} & 0.473 \\
    Generator alone & 0.477 & \best{0.500} & 0.478 & 0.444 & 0.410 \\
    Brain-Diffuser & \best{0.492} & 0.465 & 0.424 & 0.388 & 0.358 \\
    MindEye & \best{0.494} & 0.469 & 0.416 & 0.373 & 0.340 \\
    Brain-Diffuser + \ours{} layout & 0.434 & 0.484 & \best{0.499} & 0.491 & 0.474 \\
    Fixed template (control) & 0.391 & 0.414 & 0.437 & \best{0.442} & 0.435 \\
    \midrule
    \multicolumn{6}{@{}l}{\emph{Chamfer (m)\,\down}} \\
    \ours{} & 0.491 & 0.426 & \best{0.405} & 0.436 & 0.518 \\
    Generator alone & 0.426 & \best{0.391} & 0.449 & 0.597 & 0.843 \\
    Brain-Diffuser & \best{0.389} & 0.531 & 0.865 & 1.356 & 1.907 \\
    MindEye & \best{0.381} & 0.472 & 0.744 & 1.186 & 1.757 \\
    Brain-Diffuser + \ours{} layout & 0.444 & \best{0.377} & \best{0.377} & 0.439 & 0.557 \\
    Fixed template (control) & 0.524 & 0.473 & 0.437 & \best{0.424} & 0.434 \\
    \bottomrule
  \end{tabular*}

  \vspace{8pt}
  \setlength{\tabcolsep}{4pt}
  \begin{tabular}{@{}lrrll@{}}
    \multicolumn{5}{@{}l}{\textit{(b) Precision and recall at 25\,cm}} \\
    \toprule
    Method & \flhd{Precision\\\up} & \flhd{Recall\\\up} & \multicolumn{1}{c}{\flhd{$\Delta$Precision\\\up\ [95\% CI]}} & \multicolumn{1}{c}{\flhd{$\Delta$Recall\\\up\ [95\% CI]}} \\
    \midrule
    \multicolumn{5}{@{}l}{\emph{Range-calibrated, $a=2$}} \\
    \ours{} & \best{0.417} & 0.602 & \multicolumn{1}{c}{--} & \multicolumn{1}{c}{--} \\
    Generator alone & 0.362 & \best{0.741} & $+0.055$ [0.041, 0.071] & $-0.139$ [$-0.166$, $-0.113$] \\
    Brain-Diffuser & 0.317 & 0.670 & $+0.100$ [0.085, 0.114] & $-0.068$ [$-0.105$, $-0.034$] \\
    MindEye & 0.311 & 0.654 & $+0.105$ [0.086, 0.124] & $-0.052$ [$-0.092$, $-0.017$] \\
    Brain-Diffuser + \ours{} layout & 0.405 & 0.673 & $+0.011$ [0.009, 0.014] & $-0.071$ [$-0.087$, $-0.052$] \\
    \midrule
    \multicolumn{5}{@{}l}{\emph{Best tested factor by F-score, against \ours{} at $a=2$}} \\
    Generator alone, $a=1.5$ & 0.405 & 0.682 & $+0.012$ [$-0.004$, 0.028] & $-0.080$ [$-0.109$, $-0.051$] \\
    Brain-Diffuser, $a=1$ & \multicolumn{2}{c}{\emph{as above}} & $+0.005$ [$-0.006$, 0.017] & $-0.029$ [$-0.061$, 0.004] \\
    MindEye, $a=1$ & \multicolumn{2}{c}{\emph{as above}} & $-0.004$ [$-0.014$, 0.008] & $-0.018$ [$-0.043$, 0.009] \\
    \bottomrule
  \end{tabular}
\end{table}

\begin{table}[ht]
  \centering
  \caption{\textbf{Choosing $a$ without the scored scene leaves \ours{} within noise of, or worse than, the generator and the image decoders.} For each held-out scene, $a\in\{1,1.5,2,2.5,3\}$ is chosen by the Chamfer distance or F-score of the other nine. (a) Chosen $a$, folds choosing it, and scores (bold: best over both selections). (b) \ours{} minus each method, 95\% paired scene-bootstrap interval, scenes of ten favoring \ours{}. Ten-scene means.}
  \label{tab:nested_a}
  \footnotesize
  \def\flhd#1{\begin{tabular}[b]{@{}c@{}}#1\end{tabular}}
  \setlength{\tabcolsep}{2pt}
  \begin{tabular*}{\linewidth}{@{\extracolsep{\fill}}lcrrcrr@{}}
    \multicolumn{7}{@{}l}{\textit{(a) Selected factor and scores}} \\
    \toprule
    & \multicolumn{3}{c}{Selected by Chamfer distance} & \multicolumn{3}{c}{Selected by F-score} \\
    \cmidrule(lr){2-4}\cmidrule(l){5-7}
    Method & $a$ (folds) & \flhd{F@25cm\\\up} & \flhd{Chamfer (m)\\\down} & $a$ (folds) & \flhd{F@25cm\\\up} & \flhd{Chamfer (m)\\\down} \\
    \midrule
    \ours{} & 2 (10) & 0.484 & 0.405 & 2 (5), 2.5 (5) & 0.477 & 0.439 \\
    Generator alone & 1.5 (10) & \best{0.500} & 0.391 & \multicolumn{3}{c}{\emph{as by Chamfer distance}} \\
    Brain-Diffuser & 1 (10) & 0.492 & 0.389 & \multicolumn{3}{c}{\emph{as above}} \\
    MindEye & 1 (10) & 0.494 & 0.381 & \multicolumn{3}{c}{\emph{as above}} \\
    Brain-Diffuser + \ours{} layout & 1.5 (6), 2 (4) & 0.483 & 0.393 & 2 (10) & 0.499 & \best{0.377} \\
    Fixed template & 2.5 (10) & 0.442 & 0.424 & \multicolumn{3}{c}{\emph{as above}} \\
    \bottomrule
  \end{tabular*}

  \vspace{8pt}
  \setlength{\tabcolsep}{2pt}
  \begin{tabular*}{\linewidth}{@{\extracolsep{\fill}}llclc@{}}
    \multicolumn{5}{@{}l}{\textit{(b) Paired differences, \ours{} minus the method}} \\
    \toprule
    & \multicolumn{2}{c}{Selected by Chamfer distance} & \multicolumn{2}{c}{Selected by F-score} \\
    \cmidrule(lr){2-3}\cmidrule(l){4-5}
    Method & \multicolumn{1}{c}{$\Delta$ [95\% CI]} & \flhd{Scenes\\(of 10)} & \multicolumn{1}{c}{$\Delta$ [95\% CI]} & \flhd{Scenes\\(of 10)} \\
    \midrule
    \multicolumn{5}{@{}l}{\emph{$\Delta$F@25cm\,\up}} \\
    Generator alone & $-0.016$ [$-0.036$, 0.004] & 3 & $-0.023$ [$-0.041$, $-0.003$] & 2 \\
    Brain-Diffuser & $-0.008$ [$-0.027$, 0.011] & 4 & $-0.015$ [$-0.031$, 0.001] & 4 \\
    MindEye & $-0.010$ [$-0.023$, 0.004] & 4 & $-0.017$ [$-0.029$, $-0.005$] & 2 \\
    Brain-Diffuser + \ours{} layout & $+0.001$ [$-0.009$, 0.011] & 5 & $-0.021$ [$-0.027$, $-0.016$] & 0 \\
    Fixed template & $+0.042$ [0.025, 0.057] & 9 & $+0.035$ [0.020, 0.049] & 9 \\
    \midrule
    \multicolumn{5}{@{}l}{\emph{$\Delta$Chamfer (m)\,\down}} \\
    Generator alone & $+0.014$ [$-0.009$, 0.036] & 3 & $+0.048$ [0.010, 0.093] & 2 \\
    Brain-Diffuser & $+0.016$ [$-0.006$, 0.039] & 3 & $+0.050$ [0.023, 0.086] & 0 \\
    MindEye & $+0.024$ [0.008, 0.042] & 1 & $+0.058$ [0.031, 0.096] & 0 \\
    Brain-Diffuser + \ours{} layout & $+0.012$ [0.001, 0.022] & 3 & $+0.062$ [0.047, 0.078] & 0 \\
    Fixed template & $-0.019$ [$-0.040$, 0.002] & 7 & $+0.015$ [$-0.012$, 0.045] & 4 \\
    \bottomrule
  \end{tabular*}
\end{table}

\section{Loss correction and penalty ablation}
\label{app:corrected_pilot}
An earlier implementation compared the two sides of the penalty of \cref{eq:flow} in different units.

\paragraph{The correction.}
\label{app:loss_audit}
The penalty reads the layout code of a generated depth map $q$ through a fixed affine inverse of the renderer, $\mathcal R^+_{\rm aff}(q)=y_*+J^+\operatorname{vec}(q-q_*)$ with $q_*=\mathcal R(y_*)$. Here $y_*$ is the mean training layout code and $J$ a central-difference Jacobian of $\mathcal R$ at $y_*$, mean eye height and zero pitch.
This inverse returns the code in its original units, but the posterior is standardized. The correction standardizes the inverse like the posterior mean of each sampled recording, giving the map $\mathcal A_i$ of \cref{app:flow_terms}. We call checkpoints trained before this change \emph{historical} and the retrained ones \emph{corrected}. The historical generator without brain input received a zero code, which need not render as the mean training layout.

\paragraph{Effect on the results.}
All ten folds were retrained with everything else unchanged. Predictions were frozen before scoring, and no checkpoint or factor was chosen with the new scores. The corrected generator has a lower layout $r$ (0.303 against 0.369) but a lower pixel MAE and better surfaces at $a=1$. The \ours{} layout stays close to its historical value (0.480 against 0.486). \Cref{tab:surface} lists the historical and corrected rows side by side. Range calibration raises every pixel MAE. The main paper uses the corrected models throughout. An independent replay reproduced surface scores to within $4.75\times10^{-5}$ in F-score and $2.71\times10^{-6}$\,m in Chamfer distance.

\subsection{Penalty ablation}
\label{app:no_aux_ablation}
To test whether the penalty of \cref{eq:flow} helps, we retrained the corrected generators of all ten folds without it ($\lambda=0$) and compared them with $\lambda=0.1$. Both versions share seed, data, architecture and evaluation. The primary measure, fixed in advance, was the Chamfer distance of the generator at $a=1$. Without the penalty it rose by 0.1032\,m [0.0710, 0.1399], in all ten scenes, and its F-score changed by $-0.065$ [$-0.087$, $-0.045$]. The surfaces of \ours{} also worsened, with F-score changes of $-0.021$ ($a=1$) and $-0.022$ ($a=2$) in all ten scenes. The generator's F-score at $a=2$ changed within noise.

The penalty does not help every measure. Without it, the generator's layout $r$ rose by 0.1172 (from 0.301 to 0.418). Its range ratio, the predicted over the true spread of cell log depths (\cref{app:dream_adapter}), fell from 0.7835 to 0.6501 at $a=1$, farther from the ideal of one, and from 1.567 to 1.300 at $a=2$, closer to one. Its pixel MAE at $a=2$, which scores the depth map before rendering, fell by 0.283\,m. The penalty thus trades the generator's layout $r$ for better surfaces. This result rests on one training seed (all 18 measures with intervals omitted; \cref{app:extended}).

\section{Post hoc robustness analyses}
\label{app:posthoc}
All analyses in this section are post hoc. We designed them after every result above was known and ran them on the same ten scenes. Before running any of them, we wrote down six primary tests with their decision rules and all secondary quantities. Only ridge heads and ridge readouts were refit, so the generator, the readout inputs and every number elsewhere in the paper are unchanged. Independent code recomputed each primary result, and we report every outcome, favorable or not. Statistics follow \cref{app:stats} (ten scenes as units, 95\% paired scene-bootstrap intervals with 20{,}000 resamples, exact sign-flip tests over 1{,}024 sign patterns). The six primary tests are Holm-adjusted together (\cref{tab:posthoc_primary}). Steps added after a result was seen are marked. \Cref{app:extended} summarizes all secondary intervals.

\paragraph{What we had seen before the plan was frozen.}
Two results revisited here were known before we froze the plan. The first is the single-recording gain of \ours{} over the trajectory head in layout $r$ ($+0.016$, eight scenes). The second is the surface scores of \ours{} at four distance thresholds for one scene. No primary contrast had been seen, including feature replacement with single recordings.

\begin{table}[ht]
\centering
\caption{\textbf{Post hoc primary tests: two support a hypothesis, three are null and one goes against it.} Estimate: ten-scene mean and 95\% paired scene-bootstrap interval. Scenes: scenes of ten with a positive value. $p$: exact sign-flip. Holm: adjusted over the six tests. Outcome: the decision rule written before the analysis. $^\ast$A time-misaligned brain component gives the same slope, so this test does not show brain signal (text). $^\dagger$The hypothesis predicted a negative value.}
\label{tab:posthoc_primary}
\fontsize{8}{9.6}\selectfont
\setlength{\tabcolsep}{2.4pt}
\begin{tabular}{@{}>{\raggedright\arraybackslash}p{2.95cm}>{\raggedright\arraybackslash}p{3.3cm}rcrrl@{}}
\toprule
Question & Statistic & Estimate [95\% CI] & Scenes & $p$ & Holm & Outcome \\
\midrule
Does the brain's gain survive single recordings? & Layout $r$, \ours{} minus feature replacement, one recording, no pooling & $+0.012$ [0.005, 0.018] & 8 & 0.010 & 0.049 & support \\
Does averaging more recordings raise layout $r$? & Slope per doubling of recordings (1, 2, 3, 6) & $+0.0047$ [0.0038, 0.0055] & 10 & 0.002 & 0.012 & support$^\ast$ \\
Is the gain larger where the path covers less of the view? & Within-scene Spearman of path coverage and per-view gain & $+0.042$ [$-0.021$, 0.110] & 7 & 0.27 & 0.82 & null$^\dagger$ \\
Does the gain over the path grow with path noise? & Slope of \ours{} minus trajectory head over six noise levels & $-0.0006$ [$-0.0047$, 0.0036] & 4 & 0.81 & 0.82 & null \\
Is \ours{} more scene-specific than MindEye? & Correct minus wrong-scene F@25cm, \ours{} minus unmodified MindEye & $+0.014$ [$-0.009$, 0.042] & 6 & 0.37 & 0.82 & null \\
Does \ours{} gain on MindEye at coarser thresholds? & Slope of the F-score difference per doubling of the threshold (10\,cm to 1\,m) & $-0.0053$ [$-0.0079$, $-0.0024$] & 2 & 0.010 & 0.049 & contrary \\
\bottomrule
\end{tabular}
\end{table}

\paragraph{Single recordings without pooling.}
By a protocol written before this plan, we decoded each of the 416 recordings alone, with no averaging over recordings and no pooling across styles, using the models fitted on the other nine scenes. Both versions use the generator output conditioned on that recording, so only the layout differs. Replacing the EVC features with their training mean lowers the layout $r$ of \ours{} by 0.012 (primary test), in eight scenes and 277 of 416 recordings. This is about a third of the pooled effect with the control layout on the same generator (0.034). Scenes 4 and 6 are slightly negative ($-0.002$ and $-0.003$). A circular BOLD shift lowers layout $r$ by 0.016, also in eight scenes. Single-recording \ours{} is below pooled \ours{} in every scene (0.454 against 0.473). Its gain over the trajectory head is 0.016 in eight scenes, a contrast seen before the plan that scores the two with different generators. With both on one generator, the gain is 0.018 in nine scenes, and 17 of 20 participants have a positive mean gain (descriptive, as participants are nested in scenes). The readout with a \threegrid{} target does not beat its motion-alone version on single recordings ($-0.005$, four scenes), so its advantage in \cref{tab:contrasts} appears only with averaged recordings. The F-score of single-recording \ours{} is not lower than that of pooled \ours{} ($+0.008$, six scenes). These are single recordings at test time, not single-participant models, because the fitted models include the same participants' other scenes.

\paragraph{More recordings of the same videos.}
Averaging $k=1$, 2, 3 or 6 recordings per video before fusion (up to 30 random subsets per $k$) raises layout $r$ from 0.456 to 0.463, 0.465 and 0.468. This slope of 0.0047 per doubling holds in all ten scenes and meets the support rule. In a check added after the result, a 37\,s circular shift of each recording's brain component, breaking its alignment with the video, left the slope unchanged (difference $-0.00003$, five scenes). The rise thus shows only that single recordings are noisy, and the brain signal is in the levels. The gain over feature replacement is 0.013 at $k=1$ (eight scenes), 0.025 at $k=6$ and 0.026 with all recordings (ten scenes each), and at $k=6$ the intact component beats the shifted one by 0.036 in all ten scenes. A fit of $c_0-c_1/k$ gives an infinite-recording layout $r$ of 0.470. All recordings reach 0.468, and pooled \ours{} (0.473) is at this limit ($-0.003$), so more recordings of the same videos would add about 0.002. The extrapolated gain over the trajectory head, 0.032, is that reported in \cref{sec:complementary}. The brain component is reproducible across participants (split-half correlation within a video 0.61). For the readout with a \threegrid{} target, the extrapolated limit is 0.438, equal to the trajectory head ($+0.000$). These limits describe this readout and fusion head, both trained on averaged recordings, not the information available in EVC.

\paragraph{Fixed design settings and acquisition blocks.}
To test whether choosing the fixed settings on these scenes created the brain's gain, we refitted the trajectory and fusion heads, with the same readout inputs and generator, for 300 combinations of the trajectory window $K$ (4, 8, 16, 37 and 73\,s), the trajectory penalty (10 to $10^4$), the fusion penalty (1 to $10^4$) and the low-pass width $\sigma$ (2, 4 and 8 pixels) (\cref{app:extended}). \ours{} beats the trajectory head in all 300 settings, by 0.010 to 0.064 (median 0.039), in at least eight scenes in 293 settings. The other seven settings have seven positive scenes, and six of them combine $K=73$\,s with a penalty of $10^4$. The gain over feature replacement ranges from 0.030 to 0.052 (median 0.041), in at least nine scenes in all 300 settings. The paper's setting does not favor the brain, because its gains (0.035 and 0.033) are at or below these medians. It does favor both heads' absolute scores. Only 16 of the 300 settings reach its \ours{} layout $r$ of 0.480 (median 0.457, range 0.422 to 0.488), and the trajectory head is also near its best (0.446 against a median of 0.415). The paper's absolute layout $r$ values are therefore optimistic by 0.02 to 0.03 for both heads, but their difference is not. \ours{} stays above linear depth alone (0.429, not re-tuned) in 273 of 300 settings, and all 27 failures have $\sigma=8$.
Split into the three acquisition blocks (scenes 1 to 4, 5 to 7 and 8 to 10), both brain-input controls and the gain over the path are positive in every block. This follows from their being positive in all ten scenes, so it is not separate evidence, and the blocks share most participants. The one block contrast that could fail, \ours{} minus linear depth alone, is $+0.094$ and $+0.035$ in the first two blocks but $-0.011$ in scenes 8 to 10 (one of three scenes). Refitting both heads without the held-out block keeps the gain over the trajectory head at 0.041, in all ten scenes (readout and generator not refitted). Over 55 settings of voxel count, penalty, lag window and ensembles, the readout with a \threegrid{} target stays above its motion-alone version by 0.079 to 0.102 in at least nine scenes, but it is never above the trajectory head (best 0.442).

\paragraph{Where the brain's gain arises.}
To test whether the brain helps most where the camera path says little, we define the path coverage of a view as the number of its 12 azimuth bins of 7.5\textdegree{} that the camera visits within $\pm4$\,s, the window of the trajectory head. The primary test found no negative within-scene association between coverage and the per-view gain of \ours{} over feature replacement (mean Spearman correlation $+0.042$, negative in only three scenes). Descriptively, the gain is 0.013 in views whose path covers no in-view bin (five scenes positive) and 0.043 in the others (ten scenes). The gain does not fill in where the path is blind.

\paragraph{Noisy camera paths.}
At test time we added independent per-second jitter to horizontal position and heading, at five levels from 5\,cm and 1\textdegree{} to 1\,m and 20\textdegree{} (10 seeds each). For scale, the clean camera moves 0.33\,m and turns 14\textdegree{} per second on average. The jitter enters the trajectory features, the readout's camera-motion regressors and the fusion head, all trained on clean paths. Pooling partners and masks come from the clean path. The primary test, that the gain over the trajectory head grows with noise, was not supported (slope $-0.0006$, $p=0.81$). The gain rises from 0.035 to 0.063 at 0.5\,m and 10\textdegree{} (ten scenes at each level) and falls to 0.013 at 1\,m and 20\textdegree{} (seven scenes). \ours{} itself is not robust. Its layout $r$ falls from 0.473 to 0.468, 0.460, 0.419, 0.284 and 0.083, about as fast as that of the trajectory head (0.438 to 0.070). From 0.5\,m and 10\textdegree{} on, \ours{} is below the readout with a \threegrid{} target, which keeps 0.393 and 0.240 (differences $-0.109$ and $-0.157$, no scene favoring \ours{}), and below a version of itself without the path input (0.405). The fusion head relies on the path, so a path this noisy is worse than none. The gain over feature replacement stays positive at every level (0.025 to 0.053, ten scenes each), as expected, because the brain term is the same at every level. The generator detail was not resampled. \Cref{app:extended} summarizes these results and the checks with re-matched pooling partners, cumulative drift and heads refitted on noisy paths.

\paragraph{Scene-specific surface scores and distance thresholds.}
To test whether removing the wrong-scene score of \cref{tab:floors} changes the surface comparison, we reproduced every archived score within $10^{-6}$ and scored each method against its wrong-scene predictions (\cref{app:floors}) at 10, 25, 50 and 100\,cm. The scene-specific gain, correct minus wrong-scene F-score, is 0.069 for \ours{} and 0.055 for MindEye. Their difference, $+0.014$ in six scenes, comes mostly from scenes 4 and 6, and the primary test is null. It also depends on range calibration. Without it, the gain of \ours{} is below that of MindEye ($-0.018$, four scenes, added after the result). The F-score of \ours{} minus that of MindEye is $+0.005$ at 10\,cm (seven scenes), $-0.010$ at 25\,cm (four), $-0.020$ at 50\,cm (one) and $-0.011$ at 1\,m (three). The primary test, that \ours{} gains on MindEye at coarser thresholds, therefore went against the hypothesis (slope $-0.0053$ per doubling, Holm-adjusted $p=0.049$), with the deficit appearing between 10 and 25\,cm. Against Brain-Diffuser, the differences are $+0.006$, $-0.008$, $-0.014$ and $-0.007$, and no difference in scene-specific gain exceeds noise. Within our pipeline, the layout makes the surfaces more scene-specific than the generator alone calibrated in the same way ($+0.034$ at 25\,cm, eight scenes). Of the methods scored here, Brain-Diffuser + \ours{} layout has the highest mean F-score at 25\,cm, 50\,cm and 1\,m (0.499, 0.757 and 0.930), against 0.494, 0.752 and 0.927 for MindEye, a difference within noise at 25\,cm (\cref{app:layout_guidance}). No wrong-scene-corrected or threshold score shows the surfaces of \ours{} beating MindEye beyond noise.

\paragraph{What these analyses do not show.}
They reuse the development scenes, participants and scanner runs, so they test robustness, not generalization to new scenes or people. One of the two supported hypotheses, the rise with more recordings, does not show brain signal. The single-recording result has Holm-adjusted $p=0.049$, close to the threshold. The design grid varies four fixed settings. The generator, pooling rule, reliability masks, \threegrid{} cells, fusion architecture and 22-number layout code stay as chosen in development. The path noise is synthetic, and the limit over recordings is extrapolated.

\section{Post hoc analyses: withheld scenes and low-level image features}
\label{app:posthoc_extra}
We added these two analyses last, after every result above was known. Each has a written plan with decision rules, time-stamped before any of its outcomes was computed, and each was recomputed with independent code. We report every outcome here or in \cref{app:extended} and mark every step added after a result was seen. No other number of the paper changes. \Cref{app:lowlevel} reuses the ten development scenes with the statistics of \cref{app:stats}. \Cref{app:lockbox} uses scenes withheld from fitting, selecting and tuning \ours{}.

\subsection{Scenes withheld from development}
\label{app:lockbox}
\paragraph{The first withheld scene and its prior exposure.}
The first withheld scene (W1), one more of the same design (one floor plan in four styles), was recorded in the session of scenes 5 to 7 (75 recordings from 19 of the 20 participants). No component of \ours{} was fitted, selected or tuned on it, and every script that produced \ours{} refused its data. It is not unseen, though. Before \ours{} was designed, we tested an earlier readout on it once, and the recorded decision was a pass. That readout used the same six ridge configurations, trained with other scenes, but predicted a different target (13 cosine coefficients of inverse depth). The layout code, pooling, generator and both heads were designed after that outcome, and earlier generations of our pipeline had also been run on the scene. In one style the rendered camera shakes around the recorded path, so ground truth ray-cast along that path does not match what participants saw, and the plan excluded that style from the primary analysis. Five further scenes were set aside during development. One has no validated camera path and stays excluded. The other four are W2 to W5.

\paragraph{Plan for W1.}
The plan was frozen before any prediction for W1 existed. Every component of \ours{} was retrained on all ten scenes with the paper's settings. Inner selections kept the values chosen in most development folds (a 4\,s trajectory window and penalties of $10^3$ for both heads), the generator was retrained for 20{,}000 steps, and all fields were frozen with hashes before the ground truth was rendered. Before any data of W1 were loaded, a rehearsal with two development scenes held out reproduced their published predictions within $10^{-9}$. We score 27 eight-view routes in the three primary styles (56 recordings). Three primary contrasts in layout $r$ compare \ours{} with the trajectory head (H1), the fixed template (H2) and feature replacement with the generator resampled as in the paper (H3). With one scene, the unit of inference is the participant. A cluster bootstrap over the 19 participants (5{,}000 resamples) recomputes pooling, posterior, fusion and generator detail, and one-sided $p$ values are Holm-adjusted over the three contrasts. A contrast replicates if it is positive with adjusted $p<0.05$. Route and style counts are descriptive, because routes share a floor plan and, within a video, a camera path.

\paragraph{Results in W1.}
All three primary contrasts met the replication rule (W1 row of \cref{tab:lockbox_more}). Absolute accuracy did not transfer, though. \ours{} reached a layout $r$ of 0.285, below every development scene and below the 95\% prediction interval from the ten scenes (0.299 to 0.647). Every other layout was also lower than in any development scene, including the fixed template (0.178) and the generator alone (0.149), so the scene is harder for all methods. Feature replacement lowered layout $r$ in 22 of 27 routes and in every style. The gain over the trajectory head, 0.013, is about 40\% of its development mean, positive in only 16 of 27 routes and negative in one style. The planned bootstrap varies only the participants, so it is narrow. Resampling routes instead, a descriptive step added after the result, gives intervals of $-0.012$ to $+0.039$ for H1, $+0.017$ to $+0.204$ for H2 and $+0.013$ to $+0.049$ for H3. The surfaces ($a=2$, leave-one-participant-out jackknife intervals) do not reproduce the development ordering in F-score. \ours{} has a lower F-score than the generator alone calibrated in the same way (0.442 against 0.456, $-0.014$ [$-0.027$, $-0.001$]) but a lower Chamfer distance ($-0.047$\,m [$-0.061$, $-0.033$]). Its F-score is above those of the trajectory head, feature replacement, the circular shift and the fixed template ($+0.022$, $+0.033$, $+0.032$ and $+0.068$), and its Chamfer distance is below all four. \Cref{app:extended} summarizes every arm and style of W1.

\paragraph{The shaking-camera style.}
Decoded alone without pooling across styles (9 routes, 19 recordings), \ours{} was below the trajectory head in this style ($-0.026$ [$-0.033$, $-0.018$]), a contrary result by the plan's rule. Pooling all four styles by the rule of the paper (36 routes, 75 recordings) replicated all three primary contrasts, with $+0.013$ [0.004, 0.021] for H1 and $+0.032$ [0.023, 0.040] for H3.

\paragraph{Four more withheld scenes.}
\label{app:lockbox_more}
We analyzed W2 to W5 after the results of W1 were known, but their plan was frozen with that of W1, before any prediction for W1 existed. They have the same design and were recorded in a later session from six participants, all of whom also appear in the development scenes and five of whom also appear in W1. W2, W4 and W5 have four styles with six recordings each, and W3 has one style (the only one with ground truth) and six recordings, for 78 recordings and 117 eight-view routes. No camera shakes in these videos, so every style was scored. These scenes are more exposed than W1. They were training scenes of earlier model generations in this project, the earlier readout tested once on W1 had been trained with them, and on 2026-09-15, before W1 was tested, we had confirmed that this readout works on them. As for W1, no component of \ours{} was fitted, selected or tuned on them, and every script that produced \ours{} refused their data.

\paragraph{Procedure for W2 to W5.}
We applied the models of W1 (generator, fusion stack, fixed template and both heads) without refitting. Before any data of W2 to W5 were loaded, a rehearsal reproduced every prediction for W1 bit for bit, including the generator samples. We then froze the predictions of every arm with hashes and a time stamp and read the first ground-truth depth only after the freeze. This ground truth was already stored with the data, so we enforced the order by blanking those rows and logging every file that the analysis processes opened. The logs cannot rule out a read by a separate process, but after the rehearsal every prediction was a deterministic, seeded function of the frozen models, so a look at the ground truth could change no choice. Scoring followed W1, with the bootstrap over six participants. The planned five-scene summary is the mean of the per-scene differences with an exact one-sided sign-flip $p$ over the 32 sign patterns. One deviation was written down before any data of these scenes were read. W3 has one recording per participant, so we waived a rule that requires two (it only supplies the partner recording for a control not used here) and also report a four-scene summary without W3. W3 has only one video, so pooling has no partner there. An independent reviewer reproduced every number with separate code (within $10^{-15}$) and found no departure from the plan.

\begin{table}[ht]
\centering
\caption{\textbf{Five withheld scenes: \ours{} beats the fixed template in every scene, but the brain's contribution replicates within only two.} Layout $r$, route-pooled. Differences: \ours{} minus the named layout, with 95\% participant cluster bootstrap intervals that omit variation between routes. $^\ast$Replicated (positive, Holm-adjusted $p<0.05$ over the three contrasts of the scene). Mean: mean of the five differences, positive scenes and exact one-sided sign-flip $p$ (1/32 is the smallest attainable value, and the scenes are not independent, see text). Development: ten-scene mean and range of the differences (layout $r$ of \ours{} 0.359 to 0.578).}
\label{tab:lockbox_more}
\fontsize{8}{9.6}\selectfont
\setlength{\tabcolsep}{2.4pt}
\begin{tabular}{@{}lrrlll@{}}
\toprule
Scene & \begin{tabular}[b]{@{}c@{}}Recordings\\(participants)\end{tabular} & \ours{} & H1: trajectory head & H2: fixed template & \begin{tabular}[b]{@{}l@{}}H3: feature\\replacement\end{tabular} \\
\midrule
W1 & 56 (19) & 0.285 & $+0.013$ [0.006, 0.020]$^\ast$ & $+0.107$ [0.099, 0.114]$^\ast$ & $+0.030$ [0.022, 0.036]$^\ast$ \\
W2 & 24 (6) & 0.489 & $+0.031$ [0.003, 0.045]$^\ast$ & $+0.231$ [0.202, 0.246]$^\ast$ & $+0.078$ [0.048, 0.092]$^\ast$ \\
W3 & 6 (6) & 0.476 & $+0.004$ [$-0.027$, 0.030] & $+0.211$ [0.181, 0.236]$^\ast$ & $+0.034$ [0.001, 0.061]$^\ast$ \\
W4 & 24 (6) & 0.579 & $+0.033$ [$-0.001$, 0.044] & $+0.197$ [0.163, 0.209]$^\ast$ & $+0.025$ [$-0.009$, 0.037] \\
W5 & 24 (6) & 0.534 & $+0.013$ [$-0.029$, 0.035] & $+0.195$ [0.152, 0.219]$^\ast$ & $+0.041$ [$-0.004$, 0.065] \\
\midrule
Mean, scenes, $p$ & & & $+0.019$, 5, 0.031 & $+0.188$, 5, 0.031 & $+0.042$, 5, 0.031 \\
Development & & 0.473 & $+0.035$ (0.009, 0.064) & $+0.144$ ($-0.012$, 0.307) & $+0.037$ (0.012, 0.075) \\
\bottomrule
\end{tabular}
\end{table}

\paragraph{Results in each scene.}
Absolute accuracy was in the development range in the four later scenes (\cref{tab:lockbox_more}). \ours{} reached a layout $r$ of 0.489, 0.476, 0.579 and 0.534 in W2 to W5, all inside the 95\% prediction interval from the ten scenes (0.299 to 0.647), with W4 marginally above the top of the development range (0.578). Every layout was higher than in W1, so W1 was the hardest scene for every method. H2 replicated in all five scenes. H1 replicated only in W1 and W2 (W2: adjusted $p=0.016$) and was positive but missed the rule in W3 ($p=0.43$), W4 ($p=0.060$) and W5 ($p=0.33$). H3 replicated in W1, W2 and W3 (W3: adjusted $p=0.046$) and was positive but missed the rule in W4 ($p=0.083$) and W5 ($p=0.072$). The circular shift lowered layout $r$ in all four later scenes ($+0.077$, $+0.038$, $+0.029$ and $+0.040$), with intervals that exclude zero only in W2 and W3. Every primary difference in every scene fell inside its development prediction interval. On single recordings without pooling, \ours{} beat the trajectory head in 10 of 24, 2 of 6, 10 of 24 and 11 of 24 recordings of W2 to W5 (W1: 31 of 56), so the gain over the path appears only when recordings are pooled across participants. \Cref{app:extended} summarizes route counts and two secondary results of W5 that went against the development pattern.

\paragraph{Summary over the five scenes, and the unit of inference.}
\ours{} was above the trajectory head, the fixed template and feature replacement in all five scenes, with mean differences of $+0.019$, $+0.188$ and $+0.042$ and an exact sign-flip $p$ of $1/32=0.031$ each (\cref{tab:lockbox_more}). This is the smallest $p$ that five scenes allow, so it says only that all five differences were positive, not how large they are. W3 counts as positive for H1 with $+0.004$ and a within-scene $p$ of 0.43. The plan set no significance level or multiplicity rule for this summary, and Holm adjustment over the three contrasts gives 0.094 for each. Without W3, the means are $+0.023$, $+0.182$ and $+0.043$, with four of four scenes positive and $p=1/16$ each. The circular shift gives $+0.044$ (five scenes, $p=1/32$), and the readout with a \threegrid{} target minus its motion-alone version gives $+0.105$ (four scenes, $p=0.063$). The sign-flip test treats the five scenes as independent, and they are not. W2 to W5 share six participants and a session, five of the six also appear in W1, and one frozen model scores all five scenes. This positive dependence makes $p=1/32$ too small by an unknown amount. Within each scene the unit is the participant, and with six participants the intervals of W2 to W5 are rough. By the plan, no combination of these scenes is a test of generalization across scenes.

\paragraph{Surfaces.}
At the paper's factor $a=2$, \ours{} had a higher F-score than the generator alone in W2, W4 and W5, by $+0.055$ [0.020, 0.089], $+0.047$ [0.016, 0.079] and $+0.070$ [0.056, 0.084] (jackknife intervals). It tied the generator in W3 (0.000 [$-0.035$, 0.035]) and was lower in W1 ($-0.014$ [$-0.027$, $-0.001$]). Its Chamfer distance was lower than the generator's in all five scenes, by 0.085\,m to 0.128\,m in W2 to W5. Against the trajectory head, the F-score was higher in W2 ($+0.063$ [0.038, 0.088]) and W5 ($+0.038$ [0.022, 0.054]), within noise in W4 ($+0.024$ [$-0.006$, 0.055]) and lower but within noise in W3 ($-0.013$ [$-0.047$, 0.020]), where the Chamfer distance of \ours{} was worse ($+0.058$\,m [0.012, 0.104]). Without range calibration ($a=1$) the ordering reverses: \ours{} has a lower F-score than the generator alone in all five scenes and a worse Chamfer distance (point values). The F-scores are 0.375 against 0.439 in W1, 0.424 against 0.510 in W2, 0.360 against 0.472 in W3, 0.465 against 0.529 in W4 and 0.423 against 0.484 in W5. The surface gain over the generator thus depends on the range calibration, whose factor was fixed on the development scenes.

\paragraph{What the withheld scenes do and do not show.}
With every prediction frozen before the ground truth was read, \ours{} beat the fixed template within all five withheld scenes and was above the trajectory head and feature replacement in all five. The brain's contribution as defined in advance, the replication of both H1 and H3 within one scene, held in only two (W1 and W2). The gain over the trajectory head averaged $+0.019$, about half its development mean ($+0.035$), varied between routes and styles, reversed in the shaking-camera style of W1 and appeared only after pooling across participants. The absolute layout $r$ transferred in the four later scenes but not in W1. The surface gains over the generator alone held at $a=2$ in three of the four later scenes and reversed at $a=1$ in all five. These scenes were withheld, not unseen, and the earlier readout was known to work in W2 to W5. They share participants, one model and, for W1, the scanner session of scenes 5 to 7, so they are not independent replications and say nothing about new participants.

\subsection{Low-level image features}
\label{app:lowlevel}
We test whether EVC carries layout information beyond the low-level content of the frames. For each \threegrid{} cell of the frames shown at second $t$, we computed mean luminance, root-mean-square contrast, log Gabor energy (3 scales, 4 orientations) and log frame-difference motion energy within $\pm0.5$\,s. Principal component analysis on the training scenes reduced these 135 features to 16 components. We refitted the readout with a \threegrid{} target of \cref{tab:contrasts} with four input sets and all else as in the paper (six ridge configurations with fixed penalties, leave-one-scene-out folds, and the same pooling, low-frequency replacement and generator). The sets are the 12 camera-motion regressors alone (M0), motion plus the image components (M1), motion plus the 16 EVC components (M2, the paper's readout, reproduced within $10^{-6}$) and motion plus both (M3). The primary test compares M3 with M1. Earlier exploratory analyses pointed to low-level mediation, so its direction was not unexpected.

\begin{table}[ht]
\centering
\caption{\textbf{EVC adds no measurable layout $r$ to low-level image features of the frames.} Readout with a \threegrid{} target, layout $r$ over 324 routes. Differences: ten-scene mean [95\% paired scene-bootstrap CI], scenes of ten with a positive value, exact sign-flip $p$. Secondary tests are Holm-adjusted over four. Left: readouts. Right: contrasts, not tied to the row beside them.}
\label{tab:lowlevel}
\fontsize{8}{9.6}\selectfont
\setlength{\tabcolsep}{3pt}
\begin{tabular}{@{}lr@{\hspace{6pt}}lrrrl@{}}
\toprule
Inputs & Layout $r$\,\up & Contrast & Difference [95\% CI] & Scenes & $p$ & Role \\
\cmidrule(r){1-2}\cmidrule(l){3-7}
M0: camera motion & 0.338 & M3 $-$ M1 & $+0.013$ [$-0.006$, 0.033] & 6 & 0.25 & primary: null \\
M1: + image features & 0.561 & M1 $-$ M0 & $+0.223$ [0.203, 0.244] & 10 & 0.008 (Holm) & secondary \\
M2: + EVC (paper) & 0.440 & M2 $-$ M0 & $+0.098$ [0.053, 0.152] & 9 & 0.008 (Holm) & secondary \\
M3: + image features + EVC & 0.573 & M3 $-$ M2 & $+0.138$ [0.109, 0.164] & 10 & 0.008 (Holm) & secondary \\
 & & M2 $-$ M1 & $-0.125$ [$-0.161$, $-0.083$] & 1 & 0.008 (Holm) & secondary \\
\bottomrule
\end{tabular}
\end{table}

The primary test is null (\cref{tab:lowlevel}). Adding EVC to motion and image features changes layout $r$ by $+0.013$ [$-0.006$, 0.033], positive in six scenes, $p=0.25$. Every planned sensitivity analysis agrees: $+0.013$ and $+0.012$ with 8 or 32 image components, $+0.006$ with all 135 features, $+0.010$ with inner-validated penalties, and $+0.019$ [$-0.003$, 0.042] with the image features passed through a canonical hemodynamic response and the lag window of EVC. No $p$ is below 0.15, and the layout without generator detail gives $+0.014$ ($p=0.21$). With ten scenes, an EVC-specific gain of up to about 0.03 cannot be excluded. An exploratory measure points the other way. EVC explains unique variance in the per-view shape of the \threegrid{} layout before the generator (unique cross-validated $R^2$ $+0.042$ [0.016, 0.069], seven scenes, $p=0.018$). M1 beats the paper's readout (0.561 against 0.440), partly because the image features are exact for second $t$ and BOLD lags by seconds. With hemodynamically filtered features, the model without brain data reaches 0.474, within noise of 0.440 ($+0.034$ [$-0.005$, 0.070], eight scenes, $p=0.12$, added after the result). Like depth estimated from the video frames (\cref{tab:baselines}), M1 uses the stimulus frames, which \ours{} never sees, so it is a stimulus-side reference, not a competing decoder. In this dataset, the layout information of EVC is not measurably separable from the low-level content of the frames. The features do not cover eye movements or other low-level features, and the limitation in \cref{sec:conclusion} stands.

\section{Additional analyses}
\label{app:extended}
The analyses below were run on the same data but are not included in this version for space. Each line names one and states its outcome. The interactive viewer described in the reproducibility statement shows all ten atlas routes.

\begingroup\small
\begin{itemize}[leftmargin=*, itemsep=0pt, parsep=0pt, topsep=2pt]
\item \textbf{Re-rendered depth metrics:} definitions of the re-rendered depth metrics used only by the DREAM comparison below.
\item \textbf{Image-decoder outputs:} decoded images and depth of the image decoders on the atlas views. \ours{} has the lowest mean pixel MAE on these views, but over all routes it is worse than linear depth in scenes 1 and 8 and better in scene 4.
\item \textbf{DREAM depth branch, with its endpoint definitions:} the DREAM depth branch has a better layout $r$ than our generator but worse surfaces (F-score lower in all ten scenes).
\item \textbf{All transfer scores:} at $a=1$ the \ours{} layout lowers the F-score of Brain-Diffuser, MindEye and CLIP retrieval.
\item \textbf{All rows of the generator-detail ablation:} at $a=1$, removing the detail lowers the F-score by 0.031, and the template detail recovers part of this. At $a=2$ the detail raises pixel MAE by 0.016\,m.
\item \textbf{Uncalibrated planned combinations and contrasts:} the planned combinations reach F-scores of 0.371 to 0.434, below the calibrated full \ours{} (a comparison including the calibration), and the two with the \ours{} layout have a slightly higher layout $r$ than \ours{}. Uncalibrated \ours{} has a lower pixel MAE than the generator and trajectory head but a lower recall than the generator, with no scene favoring \ours{}.
\item \textbf{Precision and recall at $a=1$:} uncalibrated, \ours{} has a lower recall than the generator, Brain-Diffuser and MindEye, with precision within noise.
\item \textbf{Unaligned F-score thresholds:} no unaligned F-score threshold from 10 to 50\,cm separates \ours{} ($a=2$) from Brain-Diffuser or the generator ($a=1.5$). Within noise, image decoders have higher F-scores at 35 and 50\,cm.
\item \textbf{View-by-view lifting:} three routes lifted view by view. By the change in 3D error, \ours{} views agree more than those of Brain-Diffuser + \ours{} layout, but the two look alike by eye.
\item \textbf{Duration and lag:} the F-score of \ours{} rises from 0.445 over 6 views to 0.595 over 30, so no best duration can be identified. The change in 3D error grows with lag while the methods converge. At eight views \ours{} has a lower recall than the generator.
\item \textbf{Historical rows:} before the loss correction, the generator had a lower eight-view F-score at $a=1$ (0.420 against 0.469) but its views agreed more (one-second change 0.441 against 0.585\,m), and the change of \ours{} was the same (0.372\,m).
\item \textbf{Registration drift, all conditions:} registration drifts shift means by $<0.2$ percentage points and 2\,mm.
\item \textbf{Per-scene neural controls:} aligned brain input gives a higher layout $r$ and F-score in all ten scenes under both neural controls.
\item \textbf{Equal-scene-weight matched cohorts:} an equal-scene-weight variant of the matched cohorts gives a similar result.
\item \textbf{Penalty ablation:} without the penalty \ours{} has a slightly higher layout $r$ and a slightly lower pixel MAE, with the same range ratio.
\item \textbf{Per-scene uncalibrated ($a=1$) rows:} \ours{} has a lower F-score and a higher Chamfer distance than the generator in all ten scenes.
\item \textbf{Surface atlas of the ten atlas routes:} failures included, with the scores of \cref{tab:atlas_routes}. \ours{} has the lowest route F-score of three methods in seven of the ten routes.
\item \textbf{Depth atlas of scenes 5 to 10:} \ours{} has the highest route layout $r$ of three methods in scenes 5, 6 and 9, and Brain-Diffuser + \ours{} layout in scenes 7, 8 and 10.
\item \textbf{Noisy-path checks:} with re-matched pooling partners the gain over feature replacement stays positive. Under cumulative drift \ours{} falls from 0.473 to 0.412 while its gain over the trajectory head grows. Heads refitted on equally noisy paths recover most of the loss, with a gain (0.055) no larger than under test-time noise, and the refit left the training brain inputs unperturbed.
\item \textbf{Design settings and secondary intervals:} the 300 design settings; intervals of the 24 secondary contrasts of \cref{app:posthoc}.
\item \textbf{Secondary results:} a fit on $k\le6$ recordings predicts the held-out $k=12$ value within 0.0002. Much of the brain component is shared by videos with the same camera path (added after the result). Against the hypothesis, path coverage is positively associated with the gain over the circular shift and over a layout-only version (unadjusted $p=0.035$ and 0.010, neither surviving correction). Five other path measures, the gain over the trajectory head and a per-cell version show no detectable association. A second per-cell version, added after the result, points toward the hypothesis (unadjusted $p=0.047$).
\item \textbf{Withheld scene W1:} every arm and style of W1. The readout's gain in code similarity is slightly smaller (0.192 against 0.219), and \ours{} has a higher precision but a lower recall than the generator.
\item \textbf{Shaking-camera style of W1:} \ours{} beats feature replacement and the fixed template there, but not the trajectory head.
\item \textbf{W2 to W5:} route counts, and two secondary results of W5 against the development pattern (the direct readout below its camera-only version, and a negative gain in code similarity).
\item \textbf{Layout-code ablation (post hoc):} by the planned rule the wall distances, floor extents and opening shares each carry the brain's gain, the median and spread of depth do not, and no group is necessary. The gain raises the left-right correlation across views ($+0.039$, unadjusted) but not row order.
\end{itemize}
\endgroup

\end{document}